\documentclass[10pt,twocolumn,letterpaper]{article}

\usepackage[pagenumbers]{cvpr} 

\definecolor{cvprblue}{rgb}{0.21,0.49,0.74}
\usepackage[pagebackref,breaklinks,colorlinks,allcolors=cvprblue]{hyperref}
\usepackage{multirow}

\usepackage[accsupp]{axessibility}  

\usepackage{tabularx}
\usepackage{algorithm}
\usepackage{algpseudocode}

\usepackage{multirow}
\usepackage{colortbl}
\definecolor{c1}{RGB}{255,153,153} 
\definecolor{c2}{RGB}{255,204,153} 
\definecolor{c3}{RGB}{255,248,174} 

\newcommand{\sysname}{RGBAvatar}

\def\paperID{2972} 
\def\confName{CVPR}
\def\confYear{2025}

\title{\sysname: Reduced Gaussian Blendshapes for\\ Online Modeling of Head Avatars}

\author{
Linzhou Li\quad\quad
Yumeng Li\quad\quad
Yanlin Weng\quad\quad
Youyi Zheng\quad\quad
Kun Zhou\footnotemark[1] \\
State Key Lab of CAD\&CG, Zhejiang University\\
}

\begin{document}

\twocolumn[{%
\renewcommand\twocolumn[1][]{#1}%
\maketitle
\begin{center}
    \centering
    \captionsetup{type=figure}
    \includegraphics[width=0.95\textwidth]{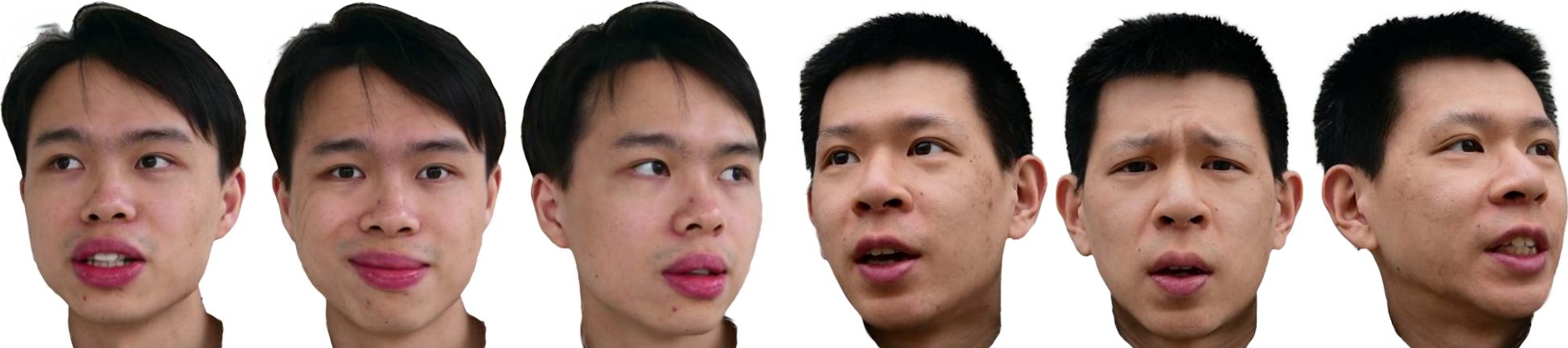}\vspace*{-2mm}
    
    \captionof{figure}{ Our \sysname~ reconstructs a high-fidelity head avatar from a 2-minute monocular video in about 80 seconds, using a reduced set of Gaussian blendshapes. These blendshapes are linearly combined to generate avatar animations in real time at about 400 FPS. }
    \label{fig:teaser}
\end{center}
}]

\begin{abstract}
We present Reduced Gaussian Blendshapes Avatar (\sysname), a method for reconstructing photorealistic, animatable head avatars at speeds sufficient for on-the-fly reconstruction. Unlike prior approaches that utilize linear bases from 3D morphable models (3DMM) to model Gaussian blendshapes, our method maps tracked 3DMM parameters into reduced blendshape weights with an MLP, leading to a compact set of blendshape bases. The learned compact base composition effectively captures essential facial details for specific individuals, and does not rely on the fixed base composition weights of 3DMM, leading to enhanced reconstruction quality and higher efficiency. To further expedite the reconstruction process, we develop a novel color initialization estimation method and a batch-parallel Gaussian rasterization process, achieving state-of-the-art quality with training throughput of about 630 images per second. Moreover, we propose a local-global sampling strategy that enables direct on-the-fly reconstruction, immediately reconstructing the model as video streams in real time while achieving quality comparable to offline settings. Our source code is available at \url{https://github.com/gapszju/RGBAvatar}.

\end{abstract}

\section{Introduction}
\label{sec:intro}

Recent advances in 3D Gaussian Splatting (3DGS) \cite{kerbl20233d} have significantly improved the reconstruction of animatable head avatars. Prior methods typically bind 3D Gaussians to parametric mesh models \cite{qian2024gaussianavatars, shao2024splattingavatar, xiang2024flashavatar, lee2024surfhead, zhao2024psavatar} or learn neutral deformation fields \cite{chen2024monogaussianavatar, xuan2024faghead, xu2024gaussian, wang2023gaussianhead}. While Gaussian Blendshapes \cite{ma20243d} have recently demonstrated superior reconstruction quality and real-time performance through efficient linear blending, they rely on predefined blendshape bases from 3D Morphable Models (3DMM) like FLAME \cite{athar2023flame} or FaceWarehouse \cite{cao2013facewarehouse}. This reliance on predefined bases results in a large number of parameters, scaling linearly with the blendshape count, leading to increased computational cost and slower training.

\renewcommand{\thefootnote}{\fnsymbol{footnote}}
\footnotetext[1]{Corresponding author (kunzhou@acm.org)}

In this paper, we introduce \sysname, a reduced Gaussian Blendshapes representation for animated head avatar reconstruction. Our key innovation is to implicitly learn a compact set of Gaussian blendshapes that effectively represent animatable head avatars. We achieve this through a lightweight MLP that maps tracked FLAME parameters to reduced blendshape weights, which are then used to linearly blend a series of learnable bases to generate an animated head avatar. By optimizing the MLP alongside the blendshapes during training, the model dynamically discovers a compact, adaptive base composition, leading to enhanced reconstruction quality and higher efficiency (both in training and runtime). 

To further expedite training, we introduce a novel color-initialization method and a batch-parallel Gaussian rasterization process. Specifically, we observe that initialized Gaussians can be treated as Gaussian kernels in 2D image space, allowing direct calculation of Gaussian properties using convolution instead of gradient descent, thus speeding up convergence with better initial estimates. Additionally, since head avatars usually require fewer than 100k Gaussians \cite{ma20243d, xiang2024flashavatar}, traditional Gaussian rasterization underutilize GPU resources. We develop a batch-parallel Gaussian rasterization to optimize GPU utilization by making the 3DGS renderer fully batch-parallel with only one GPU synchronization per training step. The acceleration enables us to achieve a training throughput of 630 images per second, reconstructing high-quality head avatars in just about 80 seconds.

With our compact representation and acceleration strategy, we significantly reduce the time required to reconstruct high-quality animatable avatars, making our method feasible for on-the-fly head avatar reconstruction. Online reconstruction allows for immediate visual feedback during data capture, which is particularly desirable in interactive applications and scenarios requiring iterative design adjustments. While our method achieves fast processing speeds, online reconstruction introduces additional challenge of sequential data arrival, thus limiting access to the entire dataset during optimization. To address this, we design a local-global sampling strategy that balances rapid adaptation to new data and rehearsal on past data, mitigating forgetting.

Experimental results demonstrate that our model captures subtle expression details using as few as 20 blendshapes and achieves nearly 400 FPS rendering speed on consumer-grade GPUs, while attaining reconstruction quality close to offline methods. Evaluations on publicly available monocular video datasets show that our approach reconstructs high-fidelity head avatar details and outperforms state-of-the-art Gaussian-based methods in both runtime and training speed.

We summarize our main contributions as follows:
\begin{itemize}
    \item We introduce Reduced Gaussian Blendshapes which efficiently captures subtle facial expressions while reducing the number of bases required to represent a head avatar.
    \item We develop a novel color-initialization method and a batch-parallel Gaussian rasterization process, significantly enhancing training throughput and optimizing GPU utilization.
    \item We establish the first on-the-fly head avatar reconstruction framework with a local-global sampling strategy, enabling real-time, online reconstruction with quality comparable to offline methods, providing immediate feedback crucial for interactive applications.
\end{itemize}

\begin{figure*}[t]
    \centering
    
    \includegraphics[width=1.0\linewidth]{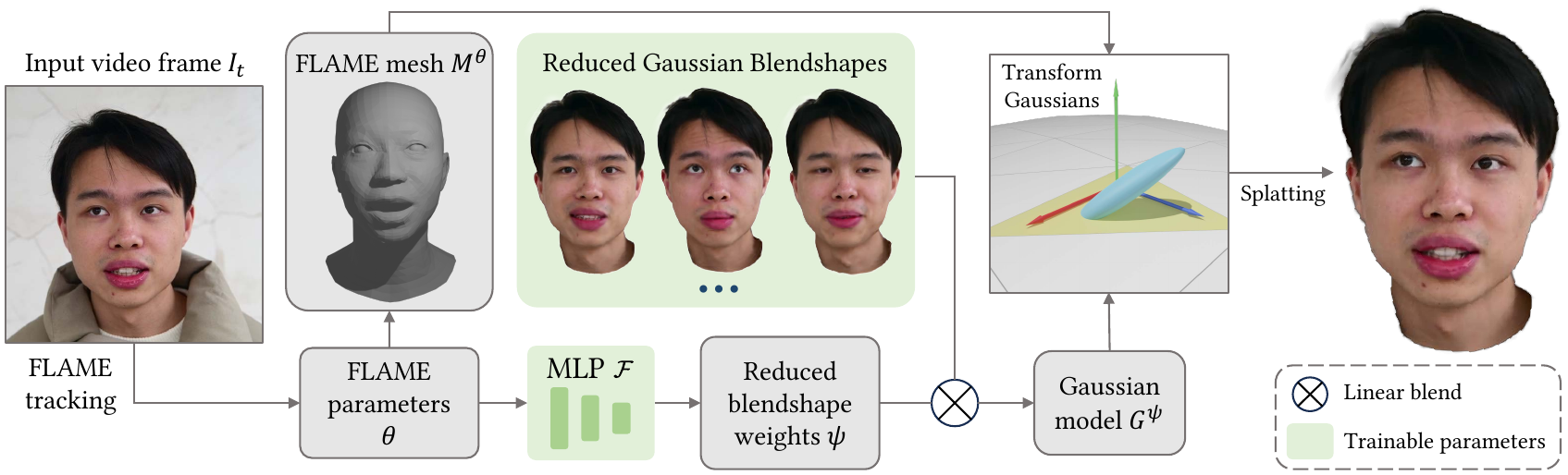}
  
    \caption{\textbf{Pipeline.} \sysname~represents the head avatar with a base model $G_0$ and a reduced set of Gaussian blendshapes $\{\Delta G_i\}_{i=1}^{K}$, each parametized as Gaussian attributes. For an input video frame $I_t$, we first track the FLAME parameters $\theta$ and generate FLAME mesh $M^{\theta}$. Then, an MLP $\mathcal{F}$ is used to map the FLAME parameters $\theta$ to the reduced blendshape weights $\psi$. The Gaussian model $G^{\psi}$ of the animated avatar is generated through linear blending with $\psi$. Finally, Gaussians are transformed into the deformed space for rendering according to the deformation of mesh triangles.
    }
    \label{fig:pipeline}
\end{figure*}

\section{Related Work}
\label{sec:related_work}

\subsection{3D Representations for Head Avatars}
Researchers have proposed various 3D representations for head avatars. The early and seminal work \cite{blanz2023morphable} proposes 3DMM to model head shape and texture in low-dimensional principal component analysis (PCA) spaces. There are many works \cite{cao2013facewarehouse, li2017learning, ploumpis2020towards, tran2018nonlinear, vlasic2006face} along this line to improve the representation ability of 3DMM. The explicit mesh is also used for reconstructing riggable head avatars from image or video inputs \cite{bai2021riggable, retsinas20243d, chaudhuri2020personalized, garrido2016reconstruction, ichim2015dynamic, khakhulin2022realistic, grassal2022neural, bharadwaj2023flare}. Recently, NHA \cite{grassal2022neural} uses two networks to predict vertex offsets and textures of underlying mesh. FLARE \cite{bharadwaj2023flare} further learns a relightable and animatable avatar through physically-based differentiable rendering.

Thanks to the success of neural radiance fields (NeRF) \cite{mildenhall2021nerf, barron2021mip, barron2022mip, barron2023zip, muller2022instant}, head avatars with implicit representation \cite{yenamandra2021i3dmm, zheng2022avatar, gafni2021dynamic, athar2022rignerf, athar2023flame, gao2022reconstructing, bai2023learning, zielonka2023instant, zhao2023havatar, gao2022reconstructing, zheng2023pointavatar, xu2023latentavatar, duan2023bakedavatar, hong2022headnerf, xu2023avatarmav} have achieved impressive results. Among them, i3DMM \cite{yenamandra2021i3dmm} presents the first deep implicit 3DMM of full heads. IMAvatar \cite{zheng2022avatar} used signed distance functions (SDF) to depict an implicit head model. HeadNeRF \cite{hong2022headnerf}  integrates NeRF to the parametric representation of the head avatar. PointAvatar \cite{zheng2023pointavatar} proposes a deformable point-based representation to overcome the limitation of explicit mesh and implicit NeRF modeling. INSTA \cite{zielonka2023instant} models a dynamic neural radiance field based on neural graphics primitives embedded around a parametric face model. It is able to reconstruct a head avatar in about ten minutes. NeRFBlendShape \cite{gao2022reconstructing} express the head model with a set of disentangled and interpretable bases, which can be driven by expression coefficients. Besides, AvatarMav \cite{xu2023avatarmav} decouples expression motion by neural voxels in a similar way.

Recently, 3DGS \cite{kerbl20233d} has made great progress in scene reconstruction, which inspires a series of works. MonoGaussianAvatar \cite{chen2024monogaussianavatar} replaces the point cloud in PointAvatar \cite{zheng2023pointavatar} with 3D Gaussians to improve rendering quality. GaussianHead \cite{wang2023gaussianhead} leverages tri-plane to store appearance related attributes. MLP based deformation field are used in \cite{wang2023gaussianhead, xuan2024faghead} for animation control. Gaussian Head Avatar \cite{xu2024gaussian} adopts a super-resolution network to achieve high-fidelity head avatar rendering. With the interpretable nature of 3DGS, some works \cite{shao2024splattingavatar, qian2024gaussianavatars, zhao2024psavatar, xiang2024flashavatar} binds the Gaussians onto the template mesh for easier head expression controllability. Among them, GaussianAvatars \cite{qian2024gaussianavatars} optimize mesh jointly for better image alignment. FlashAvatar \cite{xiang2024flashavatar} use an MLP for further adding dynamic spatial offset to Gaussians. Another line of works \cite{ma20243d, dhamo2023headgas, zielonka2024gaussian} use a group of blending bases to model the head avatar. HeadGaS \cite{dhamo2023headgas} blends learnable latent features and decode color and opacity of Gaussians from them with an MLP. GEM \cite{zielonka2024gaussian} compresses a reconstructed Gaussian avatar into eigen bases. GaussianBlendshapes \cite{ma20243d} directly blends all the Gaussian attributes and ensures the sematic consistency with mesh blendshapes, which achieves superior reconstruction quality and rendering performance. Our method follows this line of works but with a reduced blendshape model that adapts to the individual rather than maintaining the same sematic with generic 3DMM blendshapes. We demonstrate our approach delivers better reconstruction quality and higher efficiency with more compact representation.

\subsection{Real-time Facial Animation}
\citet{weise2011realtime} firstly captures facial performances in real-time by fitting a parametric blendshape model to RGB-D data based on a commodity depth sensor. Follow up works \cite{bouaziz2013online, li2013realtime, zollhofer2014real, chen2013accurate} focused on corrective shapes \cite{li2013realtime}, dynamic facial expression space \cite{bouaziz2013online} and non-rigid mesh deformation \cite{chen2013accurate, zollhofer2014real}. Although these works demonstrated impressive results, they rely on depth data which is typically unavailable in most video footage. \citet{cao20133d} presents a real-time facial animation system that fits a blendshape model from 2D video frame. DDE \cite{cao2014displaced} uses a generic regressor to get rid of calibration for each individual user. \citet{cao2015real} further regresses fine-scale face wrinkles. Face2Face \cite{thies2016face2face} captures pixel-level face performance. Our method builds upon these real-time facial animation trackers, and achieves on-the-fly photorealistic reconstruction of head avatar using 3D Gaussians.

\section{Method}




\subsection{Reduced Gaussian Blendshape Representation}
Previous blendshape modeling approaches, whether NeRF-based \cite{gao2022reconstructing} or Gaussian-based \cite{ma20243d}, are all bound to template 3DMM blendshapes. To simplify the animation control, these methods directly use the tracked 3DMM parameters to drive their avatar model, requiring a one-to-one mapping with the 3DMM blendshapes. However, traditional 3DMM models, such as FLAME \cite{athar2023flame} based on PCA or FaceWareHouse \cite{cao2013facewarehouse} based on Facial Action Coding System (FACS), use a large number of bases to capture facial expression. To balance performance and memory usage, previous methods \cite{gao2022reconstructing, ma20243d, dhamo2023headgas} typically truncate the number of blendshape bases to around 50, which constrains expressive capability.


In contrast, our reduced blendshape model leverages the full FLAME parameters, representing the head avatar with a compact set of blendshape bases that maintain both high fidelity and easy controllability. Specifically, our avatar model comprises a base Gaussian model $ G_0 $ and a set of Gaussian blendshapes $ \Delta G_1, \Delta G_2, \dots, \Delta G_K $, as shown in \cref{fig:pipeline}. Each model consists of Gaussians parameterized by position $ \mathbf{x} $, rotation $ \mathbf{q} $, scale $ \mathbf{s} $, opacity $ \alpha $, and color $ \mathbf{c} $. We generate arbitrary expressions through linear blending:
\begin{equation}
G^{\psi} = G_0 + \sum_{k=1}^{K} \psi_k \Delta G_k,
\end{equation}
where $ \psi \in \mathbb{R}^K $ represents the reduced blendshape weights. A shallow MLP $ \mathcal{F} $ maps the tracked FLAME parameters $ \theta \in \mathbb{R}^H $ (including expression coefficients and joint poses) to $ \psi $:
\begin{equation}
\psi = \mathcal{F}(\theta).
\end{equation}

To leverage geometric priors from the FLAME mesh $M^{\theta}$, we further transform the blended Gaussian model $G^{\psi}$ according to the mesh deformation, like \cite{xiang2024flashavatar}. Specifically, we define the Gaussian attributes in tangent space. For any frame $ t $, we compute the TBN (tangent, bi-tangent, normal) matrices for each mesh triangle. Using these TBN matrices and the correspondences between Gaussians and mesh triangles, we transform the Gaussians from tangent space to the deformed space for rendering.

\begin{figure}[t]
    \centering
    \includegraphics[width=1.0\linewidth]{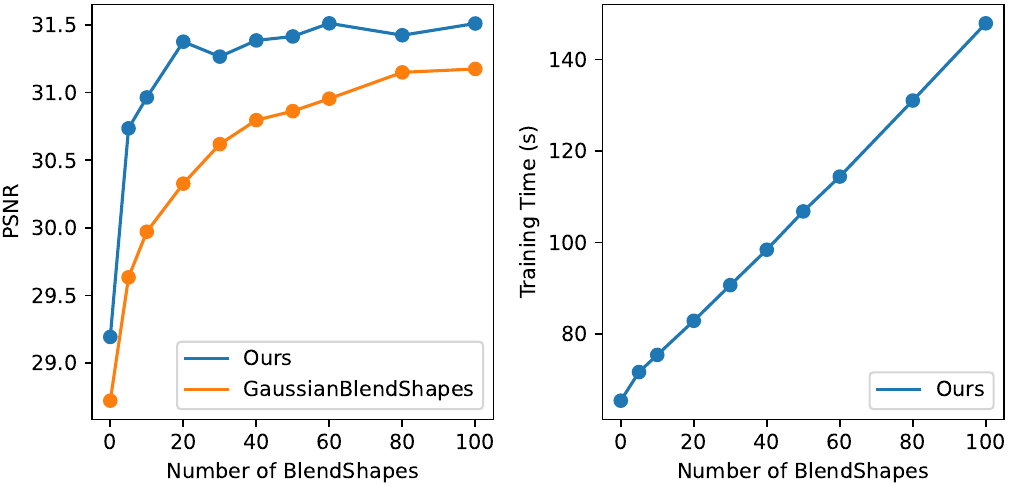}
    \caption{\textbf{Impact of number of blendshapes on reconstruction quality (left) and training time (right)}. We select 20 blendshapes to balance reconstruction quality with efficiency. Our method outperforms GaussianBlendShapes \cite{ma20243d} using fewer number of blendshapes. Experiments are conducted on INSTA \cite{zielonka2023instant} dataset.}
    \label{fig:bs_num}
\end{figure}

Our method achieves high-quality results with as few as $K=20$ blendshapes (as shown in \cref{fig:bs_num}), delivering superior runtime efficiency compared to previous works. As we adaptively reorganize arbitrary input into our reduced blendshape space, our model is also compatible with various 3DMM trackers, providing a robust, subject-adaptive avatar representation.

\subsection{Training}

\begin{figure}[t]
    \centering

    \begin{subfigure}{0.66\linewidth}
        \includegraphics[width=1.0\linewidth]{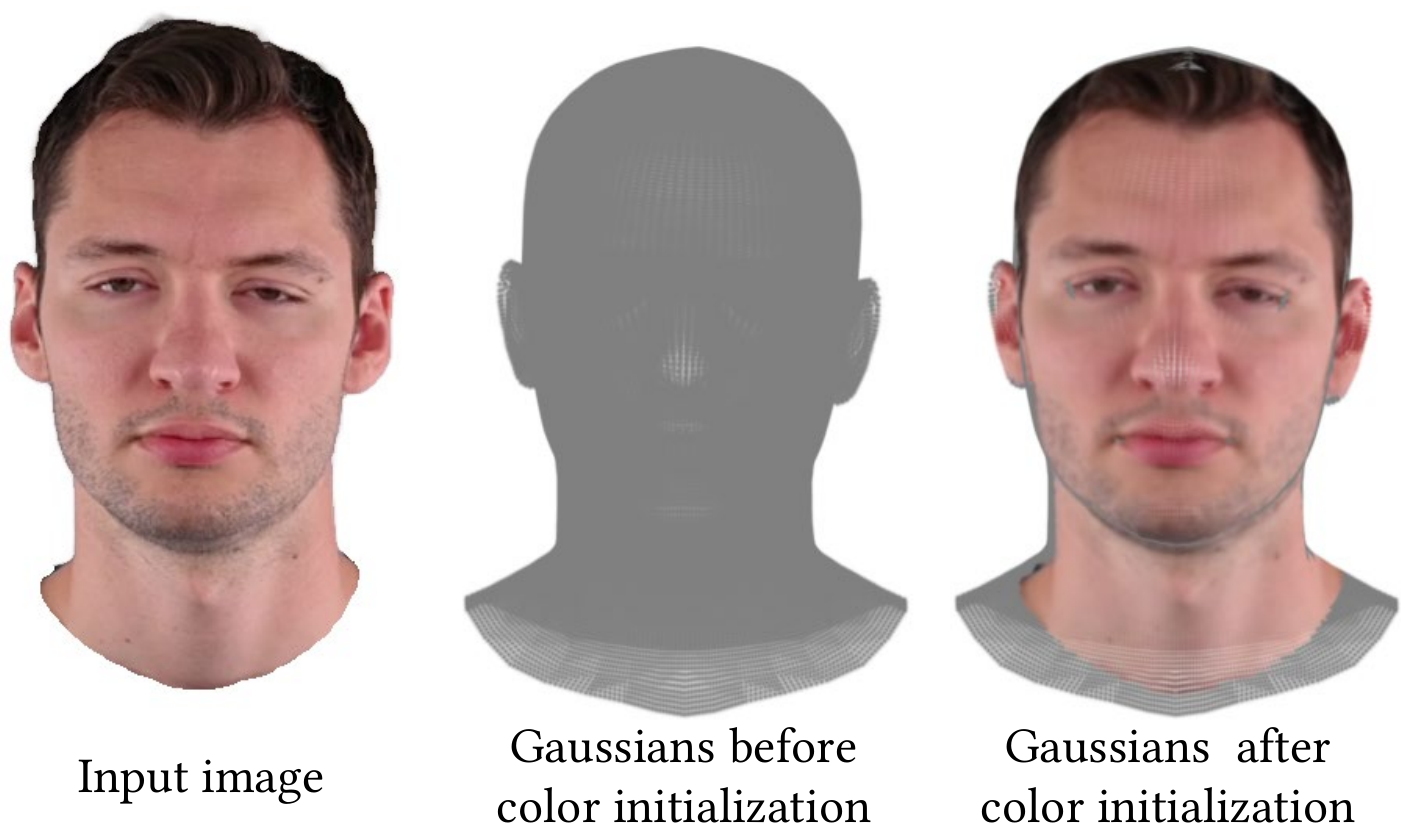}
        
    \end{subfigure}
    \hfill
    \begin{subfigure}{0.33\linewidth}
        \includegraphics[width=1.0\linewidth]{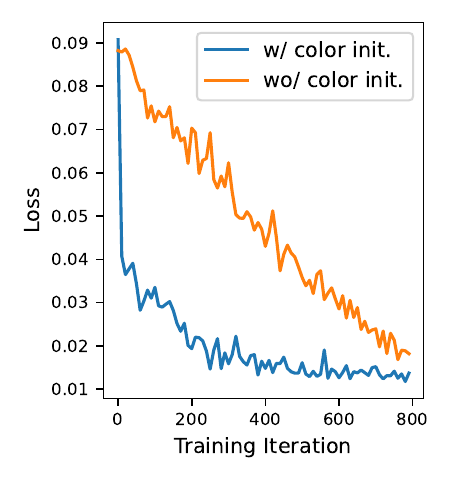}
        
    \end{subfigure}
    
    \caption{\textbf{Effect of color initialization.} This strategy is only applied once for each Gaussian during optimization. Left: results of our color initialization. Right: our color initialization strategy accelerates convergence speed in the early stage of training. }
    \label{fig:color_init}

\end{figure}

\textbf{Initialization.} To facilitate online reconstruction, we aim to sufficiently utilize the information from the 3DMM prior. Specifically, we parametrize Gaussians in tangent space of the 3DMM, with positions initialized by sampling across the UV map of the 3DMM for an even distribution over the whole mesh, similar to FlashAvatar~\cite{xiang2024flashavatar}. The rest of the Gaussian attributes are initialized with a default value for the base Gaussian model, while the Gaussian blendshapes $ \{\Delta G_i\} $ model the residual and are initialized as zero. Gaussians will remain on the same triangle they are initialized with throughout the training process.

To further accelerate convergence, we propose a \textit{color initialization} scheme that directly estimates Gaussian color attributes based on the observed pixel colors, as shown in \cref{fig:color_init}. We notice that each Gaussian can be seen as a Gaussian kernel when projected into 2D pixel space, thus a natural first estimation with their color can be done with a convolution. The color of each Gaussian is computed as:
\begin{equation}
    \mathbf{c}^{\mathrm{init}} = \frac{\sum_{i=0}^{H}\sum_{j=0}^{W} w_{ij} \mathbf{I}_{ij}}{\sum_{i=0}^{H}\sum_{j=0}^{W} w_{ij}},
\end{equation}
where $ W $ and $ H $ denote the image width and height, and $ \mathbf{I}_{ij} $ and $ w_{ij} $ are the RGB values and Gaussian splatting weights of pixel $(i, j)$, respectively. This is performed only the first time each Gaussian’s splatting weight $ w $ exceeds a threshold $ \delta = 0.1 $, skipping initialization for Gaussians that are not sufficiently visible to avoid poor estimations. This one-time initialization provides a more accurate starting value which speeds up convergence.

\noindent \textbf{Loss Function.} We apply a $ L_1 $ color loss to optimize the avatar’s reconstruction quality, with random background colors to help keep the Gaussians constrained within the head region.

\begin{figure}[t]
    \centering
    
    \includegraphics[width=1.0\linewidth]{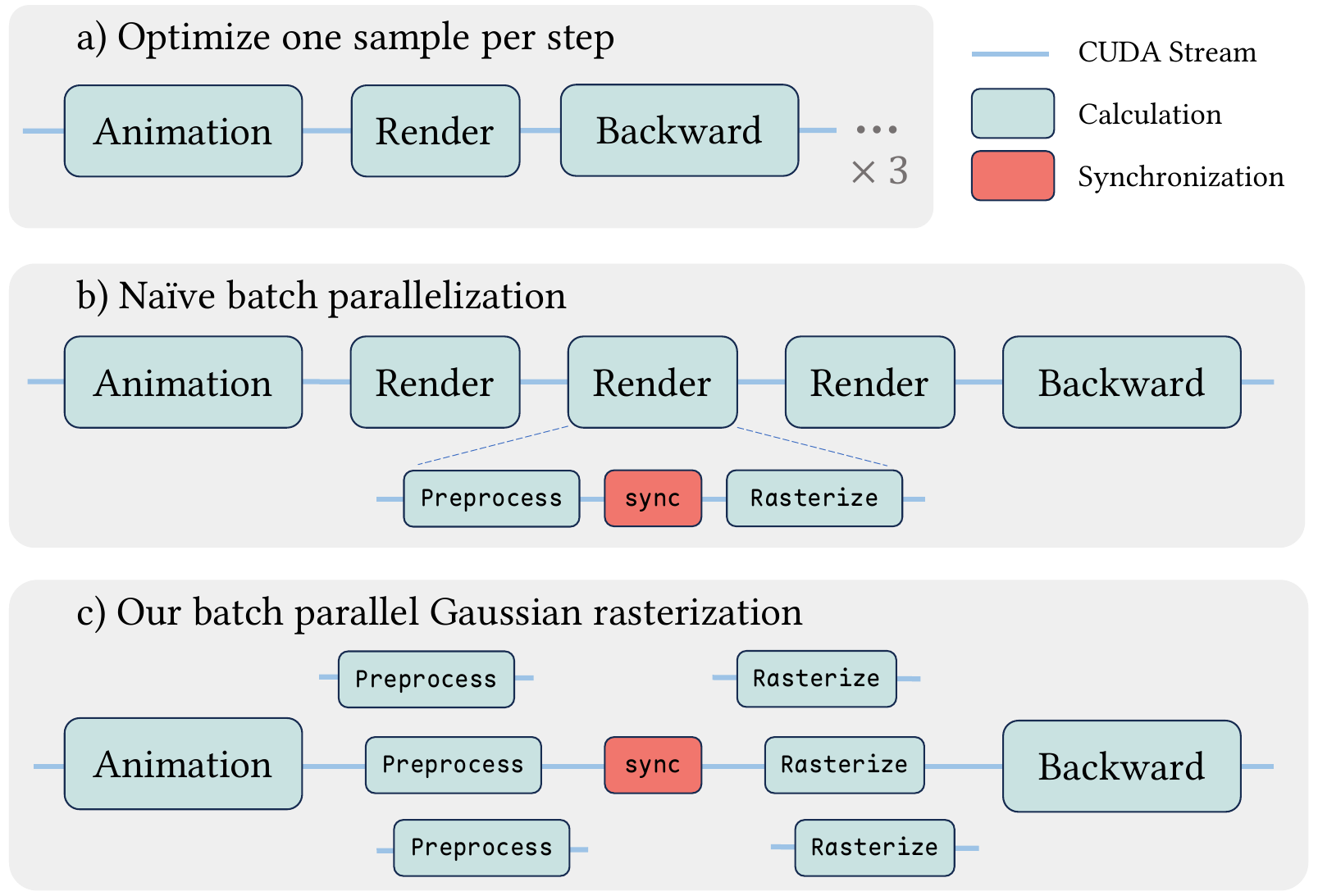}
  
    \caption{\textbf{Illustration of our batch-parallel Gaussian rasterization.} We take batch size = 3 as an example. a) Optimizing a single data sample per step, as used in \cite{xiang2024flashavatar, shao2024splattingavatar, ma20243d}, results in suboptimal GPU utilization. b) The naive batch-parallel approach, as used in \cite{chen2024monogaussianavatar}, suffers from frequent synchronization overhead. c) Our batch-parallel Gaussian rasterization mitigates synchronization issues and maximizes parallelism by leveraging CUDA streams. }
    \label{fig:batch}
\end{figure}

\noindent \textbf{Batch-parallel Gaussian Rasterization.} Previous methods \cite{xiang2024flashavatar, shao2024splattingavatar, ma20243d, qian2024gaussianavatars} often achieve suboptimal GPU utilization during training, typically below 60\%, as they follow the original 3DGS training scheme of optimizing one sample per step (see \cref{fig:batch}a). Given that head avatars typically involve fewer than 100k Gaussians, we identified that the bottleneck lies not in GPU computation but in frequent GPU-CPU synchronizations.

To address this, we optimize a batch of samples in one step. For Gaussian rendering, a naive approach is to rasterize images in a batch sequentially like \cite{chen2024monogaussianavatar}, but this method still requires GPU-CPU synchronization between \texttt{preprocess} and \texttt{rasterize} calculations (see \cref{fig:batch}b) because memory allocation for \texttt{rasterize} calculation depends on the results of \texttt{preprocess} calculation. However, theoretically only one synchronization is needed per batch, as the GPU rasterization depends on the Gaussian preprocessing outcome regardless of batch size. Thus, we split Gaussian Splatting into two stages and perform synchronization only once after \texttt{preprocess}.

Additionally, since samples in a batch are fully parallelizable in both Gaussian preprocessing and rasterization stages, we assign them to different CUDA Streams to maximize computation, as shown in \cref{fig:batch}c. With this batch-based training scheme, we achieve 100\% Stream Processor utilization during reconstruction, effectively mitigating the GPU-CPU synchronization bottleneck. Note that gsplat \cite{ye2024gsplat} also supports batched Gaussian rendering. However, it is limited to rendering the same set of Gaussians from different camera views, whereas our method allows rendering different sets of Gaussians from the same camera view.

\subsection{Online Reconstruction Process}

With our compact representation and efficient training scheme, our model is able to process input frames in a speed faster than real-time recording (25 FPS), which makes on-the-fly reconstruction possible. However, applying this to an online setting introduces challenges beyond convergence speed, particularly in managing streaming data. Unlike offline reconstruction, which has complete access to the dataset, online reconstruction operates on a continuous data stream. This setup creates two somewhat conflicting demands: (1) we need fast convergence on the newly arriving, previously unseen data to achieve \textit{quick adaptation}, and (2) we must mitigate \textit{catastrophic forgetting} of past data. To balance these goals, we introduce a \textit{global-local sampling strategy}. This strategy maintains a local sampling pool $\mathcal M_l$ for newly introduced, under-optimized frames and a sampling pool $\mathcal M_g$ for historical frames. By sampling from both pools in each batch, the gradient calculations incorporate both recent and past information, effectively balancing rapid adaptation to new data with the retention of previously learned knowledge, thereby mitigating forgetting.

\noindent \textbf{Online Optimization.} During each optimization step, we split the batch size $B$ into a local batch $B_l$ and a global batch $B_g$ according to a predefined ratio $\eta$. We then sample $B_l$ from the local memory $\mathcal M_l$ and $B_g$ from the global memory $\mathcal M_g$, optimizing the combined batch ${B_l, B_g}$ via gradient descent. In our setup, we set $\eta=0.7$, with $|\mathcal M_l|=150$ and $|\mathcal M_g|=1000$, prioritizing recent frames in the sampling process.

\noindent \textbf{Processing Incoming Frames.} For each incoming frame, we first extract the 3DMM parameters \cite{cao2014displaced} and mask the background \cite{rvm} using off-the-shelf methods. The new data sample is then added to $\mathcal M_l$ in a First-In-First-Out manner. When $\mathcal M_l$ reaches capacity, the oldest sample is removed and added to $\mathcal M_g$ using Reservoir Sampling \cite{vitter1985random}, ensuring each frame in the data stream has an equal probability of being retained in the global pool. Please refer to supplementary materials for detailed algorithm.

\begin{table*}[t]
\centering
\resizebox{\linewidth}{!}{
\begin{tabular}{|cc|cccccccccccc}
\hline
\multicolumn{2}{c|}{\multirow{2}{*}{Datasets}} & \multicolumn{8}{c|}{INSTA dataset} & \multicolumn{4}{c}{GaussianBlendShapes dataset} \\  \cline{3-14} \multicolumn{2}{c|}{} & bala & biden & justin & malte\_1 & marcel & nf\_01 & nf\_03 &\multicolumn{1}{c|} {wojtek\_1} & subject1 & subject2 & subject3 & subject4  \\ \hline

\multicolumn{1}{c|}{\multirow{8}{*}{PSNR $\uparrow$}}
& SplattingAvatar & 30.99 & 31.24 & 31.90 & 28.08 & 25.65 & 27.52 & 27.80 & \multicolumn{1}{c|}{31.48} & 31.30 & 32.52 & 31.96 & 34.69 \\
\multicolumn{1}{c|}{}
& GaussianAvatars & 31.58 & 31.20 & 31.80 & 27.60 & 25.93 & 27.55 & 28.04 & \multicolumn{1}{c|}{32.10} & 31.03 & 33.35 & 32.30 & 34.51 \\
\multicolumn{1}{c|}{}
& FlashAvatar & 31.98 & 23.76 & 31.50 & 28.41 & 23.19 & 26.89 & 26.97 & \multicolumn{1}{c|}{31.60} & 29.67 & 32.61 & 30.19 & 32.29 \\
\multicolumn{1}{c|}{}
& MonoGaussianAvatar & 28.71 & 32.56 & 32.13 & 26.37 & 23.64 & 28.48 & \cellcolor{c1}31.00 & \multicolumn{1}{c|}{31.29} & 33.72 & 33.67 & 33.13 & 33.67 \\
\multicolumn{1}{c|}{}
& GaussianBlendShapes & \cellcolor{c2}33.21 & 32.69 & 32.73 & 28.38 & 26.49 & 27.99 & 28.58 & \multicolumn{1}{c|}{32.76} & 33.14 & 33.38 & 32.30 & 34.93 \\
\multicolumn{1}{c|}{}
& Ours & \cellcolor{c1}33.89 & \cellcolor{c1}32.93 & \cellcolor{c2}32.99 & \cellcolor{c1}29.54 & \cellcolor{c1}27.72 & \cellcolor{c2}28.87 & \cellcolor{c2}30.79 & \multicolumn{1}{c|}{\cellcolor{c2}32.77} & 34.85 & \cellcolor{c2}34.53 & \cellcolor{c1}34.47 & \cellcolor{c2}35.77 \\
\multicolumn{1}{c|}{}
& Ours (w/ LPIPS) & 32.75 & \cellcolor{c2}32.80 & \cellcolor{c1}33.16 & 29.22 & \cellcolor{c2}27.57 & 28.81 & 30.74 & \multicolumn{1}{c|}{\cellcolor{c1}33.15} & \cellcolor{c1}35.08 & \cellcolor{c1}34.64 & 34.17 & \cellcolor{c1}35.96 \\
\multicolumn{1}{c|}{}
& Ours (online) & 32.94 & 32.60 & 32.79 & \cellcolor{c2}29.27 & 27.28 & \cellcolor{c1}28.89 & 30.46 & \multicolumn{1}{c|}{32.65} & \cellcolor{c2}34.92 & 33.97 & \cellcolor{c2}34.36 & 35.58 \\
\cline{1-14}

\multicolumn{1}{c|}{\multirow{8}{*}{SSIM $\uparrow$}}
& SplattingAvatar & 0.9222 & 0.9561 & 0.9606 & 0.9344 & 0.9085 & 0.9285 & 0.9179 & \multicolumn{1}{c|}{0.9505} & 0.9250 & 0.9506 & 0.9211 & 0.9615 \\
\multicolumn{1}{c|}{}
& GaussianAvatars & 0.9457 & 0.9593 & 0.9657 & 0.9424 & 0.9281 & 0.9422 & 0.9333 & \multicolumn{1}{c|}{0.9622} & 0.9338 & 0.9635 & 0.9397 & 0.9673 \\
\multicolumn{1}{c|}{}
& FlashAvatar & 0.9154 & 0.8670 & 0.9481 & 0.9290 & 0.8962 & 0.9259 & 0.9069 & \multicolumn{1}{c|}{0.9460} & 0.9115 & 0.9474 & 0.9134 & 0.9425 \\
\multicolumn{1}{c|}{}
& MonoGaussianAvatar & 0.9315 & 0.9681 & 0.9667 & 0.9417 & 0.9232 & 0.9416 & \cellcolor{c2}0.9458 & \multicolumn{1}{c|}{0.9612} & 0.9513 & 0.9677 & 0.9449 & 0.9631 \\
\multicolumn{1}{c|}{}
& GaussianBlendShapes & 0.9455 & 0.9658 & 0.9672 & 0.9455 & 0.9270 & \cellcolor{c2}0.9474 & 0.9352 & \multicolumn{1}{c|}{0.9640} & 0.9428 & 0.9607 & 0.9418 & 0.9693 \\
\multicolumn{1}{c|}{}
& Ours & \cellcolor{c1}0.9549 & \cellcolor{c1}0.9720 & \cellcolor{c1}0.9721 & \cellcolor{c1}0.9522 & \cellcolor{c1}0.9388 & \cellcolor{c1}0.9481 & \cellcolor{c1}0.9474 & \multicolumn{1}{c|}{\cellcolor{c1}0.9665} & \cellcolor{c1}0.9549 & \cellcolor{c1}0.9708 & \cellcolor{c2}0.9507 & \cellcolor{c1}0.9721 \\
\multicolumn{1}{c|}{}
& Ours (w/ LPIPS) & \cellcolor{c2}0.9498 & \cellcolor{c2}0.9713 & \cellcolor{c2}0.9715 & 0.9501 & \cellcolor{c2}0.9361 & 0.9472 & \cellcolor{c2}0.9458 & \multicolumn{1}{c|}{\cellcolor{c2}0.9660} & 0.9533 & 0.9690 & 0.9474 & 0.9706 \\
\multicolumn{1}{c|}{}
& Ours (online) & 0.9509 & 0.9699 & 0.9705 & \cellcolor{c2}0.9502 & 0.9333 & 0.9464 & 0.9430 & \multicolumn{1}{c|}{0.9655} & \cellcolor{c2}0.9553 & \cellcolor{c2}0.9699 & \cellcolor{c1}0.9518 & \cellcolor{c2}0.9718 \\
\cline{1-14}

\multicolumn{1}{c|}{\multirow{8}{*}{LPIPS $\downarrow$}}
& SplattingAvatar & 0.1160 & 0.0599 & 0.0688 & 0.0841 & 0.1654 & 0.1333 & 0.1216 & \multicolumn{1}{c|}{0.0773} & 0.1265 & 0.1022 & 0.1546 & 0.0944 \\
\multicolumn{1}{c|}{}
& GaussianAvatars & 0.0699 & 0.0538 & 0.0605 & 0.0701 & 0.1389 & 0.1153 & 0.0941 & \multicolumn{1}{c|}{0.0533} & 0.1028 & 0.0779 & 0.1264 & 0.0745 \\
\multicolumn{1}{c|}{}
& FlashAvatar & 0.0853 & 0.1518 & 0.0745 & 0.0687 & 0.1620 & 0.1124 & 0.1189 & \multicolumn{1}{c|}{0.0628} & 0.0893 & 0.0611 & 0.0969 & 0.0881 \\
\multicolumn{1}{c|}{}
& MonoGaussianAvatar & 0.0668 & \cellcolor{c2}0.0370 & \cellcolor{c2}0.0494 & 0.0591 & \cellcolor{c2}0.1241 & \cellcolor{c2}0.1025 & \cellcolor{c2}0.0811 & \multicolumn{1}{c|}{\cellcolor{c2}0.0431} & \cellcolor{c2}0.0797 & \cellcolor{c2}0.0478 & \cellcolor{c2}0.0892 & 0.0579 \\
\multicolumn{1}{c|}{}
& GaussianBlendShapes & 0.0833 & 0.0537 & 0.0647 & 0.0722 & 0.1477 & 0.1210 & 0.0992 & \multicolumn{1}{c|}{0.0589} & 0.1066 & 0.0838 & 0.1260 & 0.0705 \\
\multicolumn{1}{c|}{}
& Ours & \cellcolor{c2}0.0663 & 0.0408 & 0.0518 & \cellcolor{c2}0.0586 & 0.1249 & 0.1098 & 0.0865 & \multicolumn{1}{c|}{0.0507} & 0.0847 & 0.0613 & 0.1026 & 0.0567 \\
\multicolumn{1}{c|}{}
& Ours (w/ LPIPS) & \cellcolor{c1}0.0430 & \cellcolor{c1}0.0330 & \cellcolor{c1}0.0428 & \cellcolor{c1}0.0434 & \cellcolor{c1}0.0991 & \cellcolor{c1}0.0919 & \cellcolor{c1}0.0695 & \multicolumn{1}{c|}{\cellcolor{c1}0.0323} & \cellcolor{c1}0.0491 & \cellcolor{c1}0.0361 & \cellcolor{c1}0.0536 & \cellcolor{c1}0.0347 \\
\multicolumn{1}{c|}{}
& Ours (online) & 0.0666 & 0.0430 & 0.0534 & 0.0609 & 0.1315 & 0.1135 & 0.0906 & \multicolumn{1}{c|}{0.0529} & 0.0820 & 0.0561 & 0.0996 & \cellcolor{c2}0.0545 \\
\hline
\end{tabular}
}
\caption{\textbf{Quantitative comparisons.} We compare our approach with five Gaussian-based state-of-the-art methods. The \colorbox{c1}{best} and \colorbox{c2}{second} results are highlighted.}
\label{tab:eval_quality}
\end{table*}
\begin{table}[t]
\centering
\begin{tabular}{l|cc}
\toprule
Method & Training Time & Rendering FPS \\
\midrule
SplattingAvatar & 21min & 300 \\
GaussianAvatars & 14min & 177 \\
FlashAvatar & 17min & 304 \\
MonoGaussianAvatar & 9h & 17 \\
GaussianBlendShapes & 20min & 267 \\
Ours & \textbf{81s} & \textbf{398} \\
\bottomrule
\end{tabular}
\caption{\textbf{Performance comparisons.} We report the training time and run time FPS on RTX 3090. Note that we train GaussianAvatars~\cite{qian2024gaussianavatars} for 40k iterations for monocular setting. And the run time calculation includes both rendering and animation. }
\label{tab:eval_perf}
\end{table}



\begin{figure*}[t]
    \centering
    
    \includegraphics[width=1.0\linewidth]{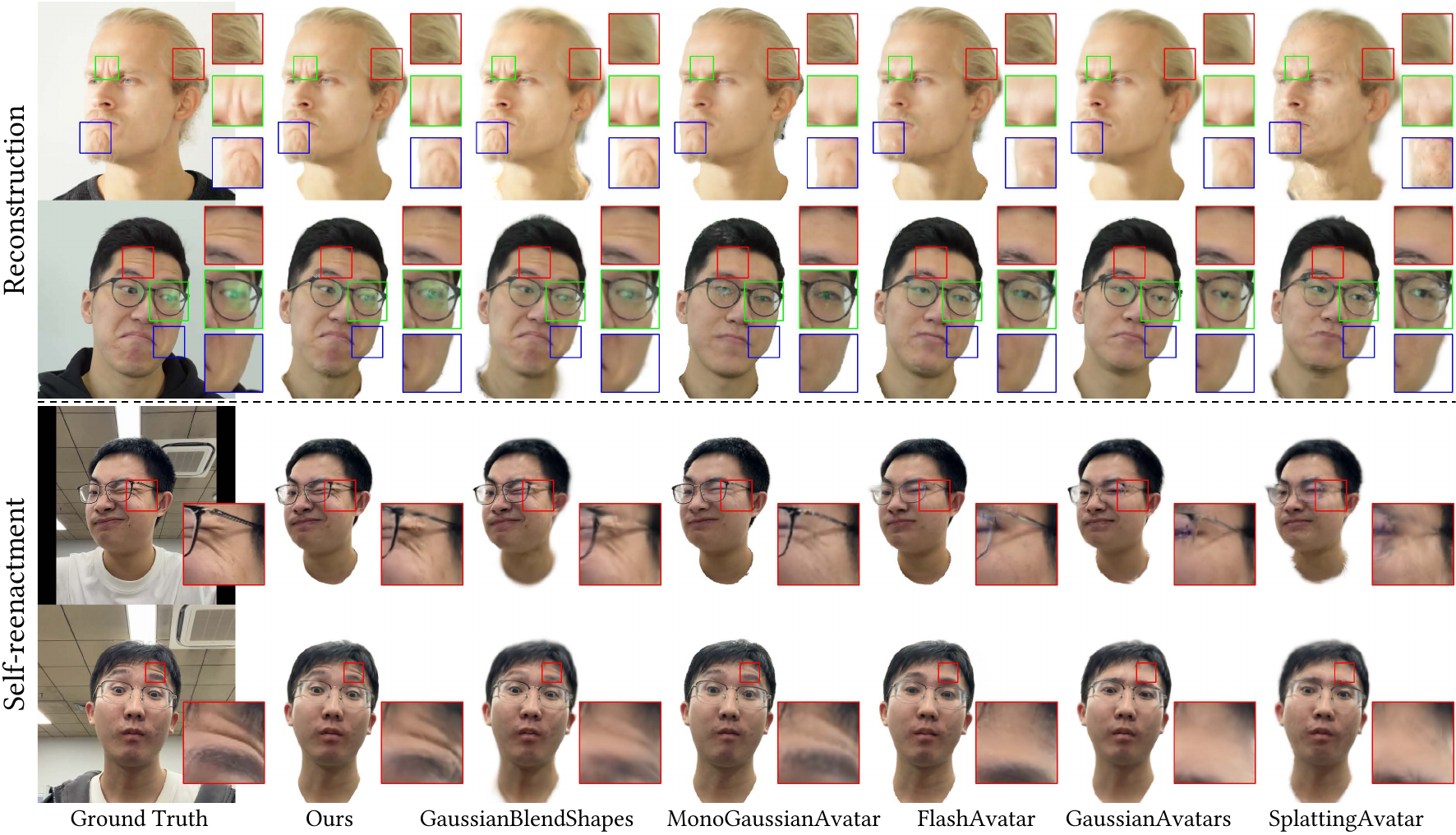}
  
    \caption{\textbf{Qualitative comparisons.} Our method produces photorealistic rendering of wrinkles, mouth, hair and eye regions. Compared to GaussianBlendShapes \cite{ma20243d}, which uses 50 blendshape bases, our approach requires only 20 bases while capturing even finer details such as reflections on eyeglasses and deep wrinkles. }
    \label{fig:compare}
\end{figure*}

\section{Experiments} 
\subsection{Setup}
\textbf{Baselines.} We compare our method against five state-of-the-art Gaussian-based head avatar reconstruction approaches: SplattingAvatar \cite{shao2024splattingavatar} and GaussianAvatars \cite{qian2024gaussianavatars}, which embed Gaussians on mesh surfaces; FlashAvatar \cite{xiang2024flashavatar}, which further models spatial offsets of 3D Gaussians using a MLP; MonoGaussianAvatar \cite{chen2024monogaussianavatar}, which learns a Gaussian deformation field for avatar animation; and GaussianBlendshapes \cite{ma20243d}, which optimizes a set of Gaussian blendshape bases driven by FLAME blendshape weights.

\noindent \textbf{Datasets.} We evaluate our method on eight videos from the INSTA Dataset \cite{zielonka2023instant} and four videos from the GaussianBlendshapes Dataset \cite{ma20243d}. All videos are cropped and resized to $512^2$ resolution, with sequence lengths between 2000 and 5000 frames. Consistent with previous work \cite{zielonka2023instant}, we reserve the last 350 frames of each video for self-reenactment testing. Backgrounds are removed using Robust Video Matting \cite{rvm}, and each frame's FLAME parameters are extracted via an offline tracker \cite{mica}. Please refer to supplementary materials for implementation details.

\subsection{Comparisons}
We evaluate our results using standard metrics in animatable avatar reconstruction, including PSNR, SSIM, and LPIPS \cite{zhang2018unreasonable}. As shown in \cref{tab:eval_quality}, our method outperforms baselines in PSNR and SSIM across most cases, particularly in challenging scenarios such as the marcel video. Note that MonoGaussianAvatar \cite{chen2024monogaussianavatar} utilizes LPIPS loss during training which leads to a better LPIPS result. We show that our method achieves superior results if we add the LPIPS loss with a weight of 0.05 in training.

To better demonstrate our effectiveness, we recorded two additional videos with expressive expressions for a qualitative evaluation. Each contains 3k frames and the last 1.5k are held out as test set. As shown in \cref{fig:compare}, our method produces photorealistic rendering of wrinkles, hair, mouth and eye regions. SplattingAvatar \cite{shao2024splattingavatar} and GaussianAvatars \cite{qian2024gaussianavatars} struggle to capture expressive facial details, as their Gaussian transformations are entirely driven by mesh deformation, limiting their expressiveness. MonoGaussianAvatar \cite{chen2024monogaussianavatar} and FlashAvatar \cite{xiang2024flashavatar} use MLPs to model Gaussian deformation but fail to capture high-frequency details, such as wrinkles, due to the smoothness bias of MLPs, even with extended training time (\eg 9h in MonoGaussianAvatar). GaussianBlendshapes \cite{ma20243d} can capture wrinkle-level details with 50 blendshape bases, but our method outperforms them in capturing finer details. Specifically, by leveraging all 100 expression coefficients, we capture large facial expressions, such as deep wrinkles (4th and 5th rows in \cref{fig:compare}). Our method can also utilize pose rotations to improve performance on side views, capturing details like hair strands (1st row), eyeglass frames (4th row), and reflections on glass (2nd row), all of which are strongly correlated with pose.

\cref{tab:eval_perf} shows the training and runtime performance comparisons. Our method reconstructs a head avatar within about 80 seconds — 400 times faster than MonoGaussianAvatar and 10 times faster than GaussianAvatars. For rendering, our method is also the most efficient, achieving 398 FPS. Note that our runtime includes both rendering and animation calculation (compute the FLAME mesh, linearly blend the Gaussian attributes, and transform Gaussians with mesh deformation).

\begin{figure}[t]
    \centering
    
    \includegraphics[width=1.0\linewidth]{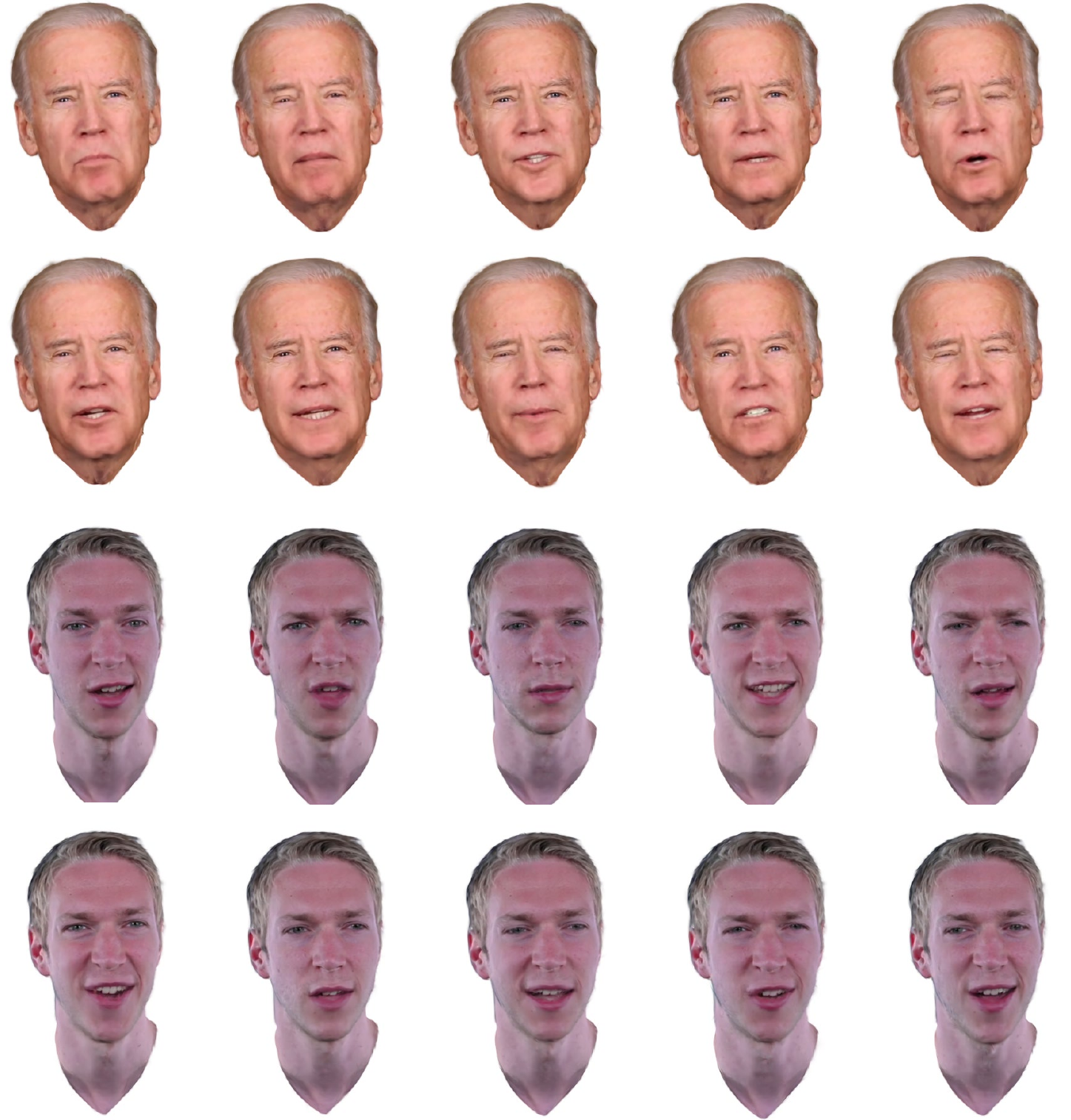}
  
    \caption{\textbf{Blendshape visualization.} We show 10 out of 20 blendshapes learned by our method for two subjects. The blendshapes are diverse and adapted to each subject.}
    \label{fig:bs_vis}
\end{figure}
\begin{figure}[t]
    \centering
    
    \includegraphics[width=1.0\linewidth]{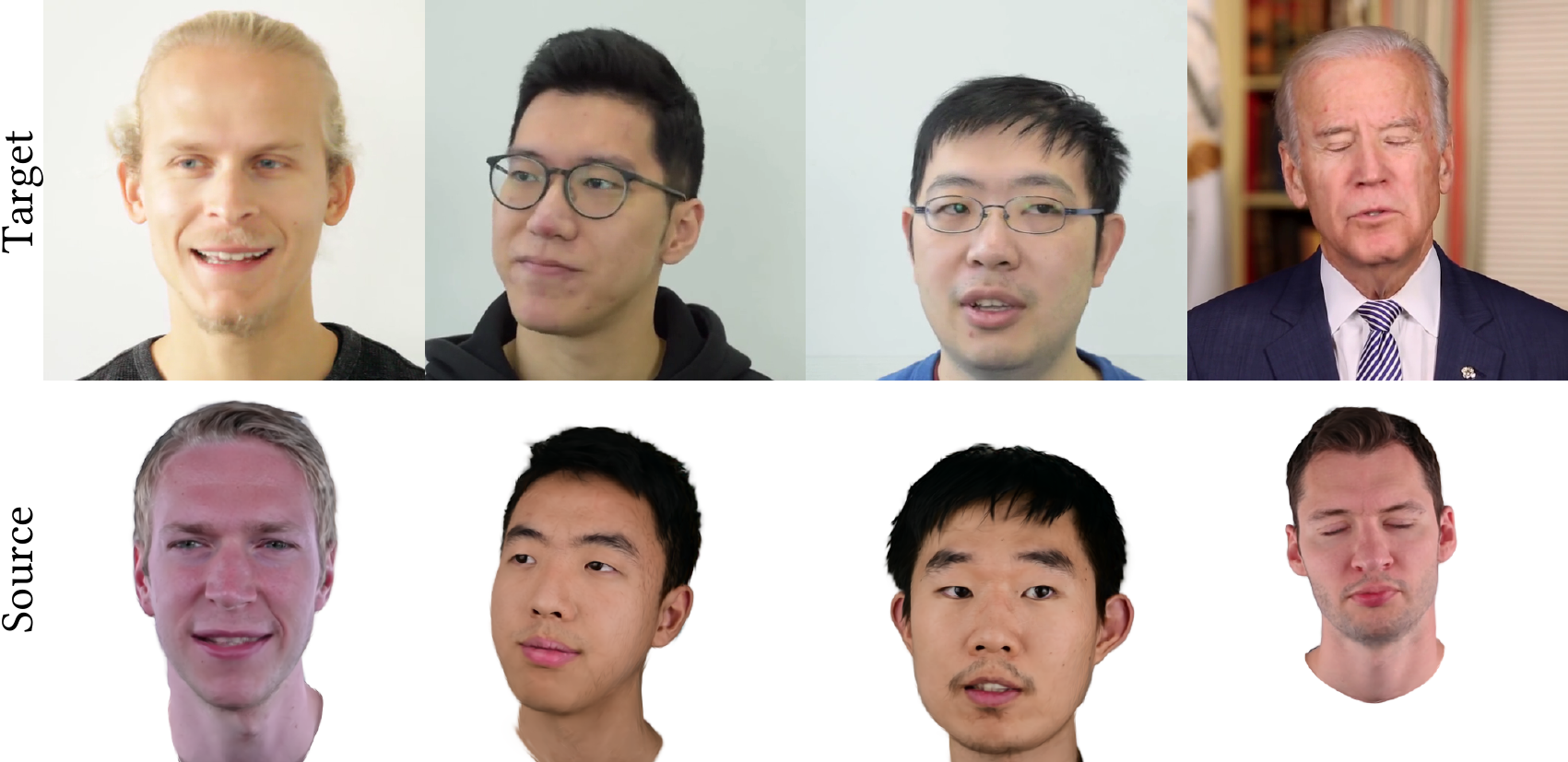}
  
    \caption{\textbf{Cross-identity reenactment.} Our reduced blendshapes can be driven by FLAME parameters easily. We achieve natural results on reenactment task. }\vspace*{-2mm}
    \label{fig:cross_id}
\end{figure}
\begin{figure}[t]
    \centering
    
    \includegraphics[width=1.0\linewidth]{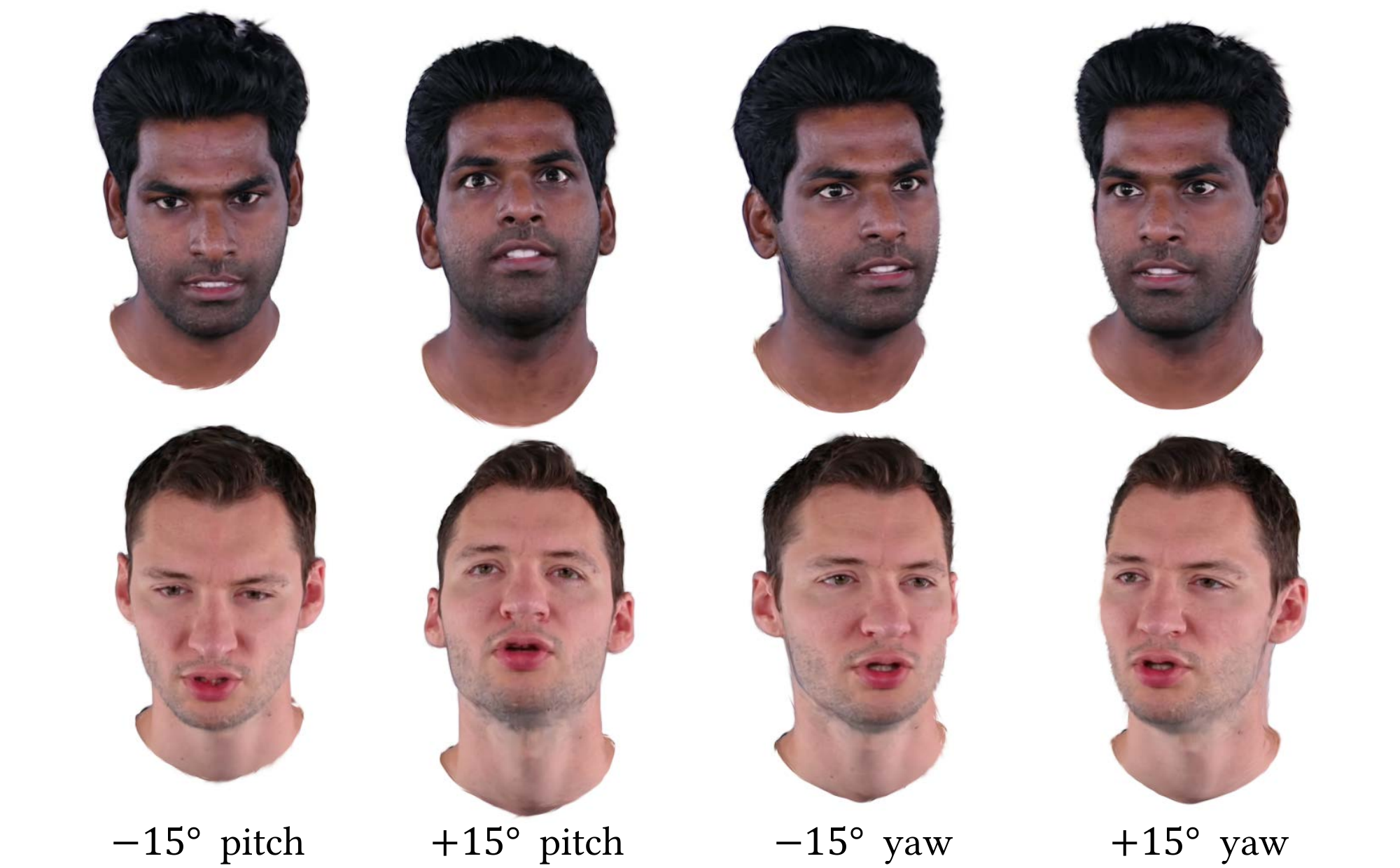}
  
    \caption{\textbf{Results of novel view extrapolation.}}
    \label{fig:novel_view}
\end{figure}

\subsection{Blendshape Visualization}
\cref{fig:bs_vis} visualizes the 10 out of 20 learned blendshapes for two subjects. We observe that each blendshape corresponds to a specific expression, with distinct semantic variations across individuals. In this experiment, we position the Gaussians in world space rather than tangent space to provide clearer visualization.

\subsection{Applications}
\cref{fig:cross_id} demonstrates cross-identity reenactment, where tracked coefficients from a target avatar (top row) are used to drive a source avatar (bottom row). Our adaptive blendshapes are fully compatible with 3DMM, allowing for straightforward exchange of expression parameters to achieve seamless facial reenactment. In \cref{fig:novel_view}, we render the avatars from four novel viewpoints, distinct from the training views. Our model consistently maintains expressions across these different perspectives.

\subsection{Ablation Studies}
\begin{figure}[t]
    \centering
    \includegraphics[width=1.0\linewidth]{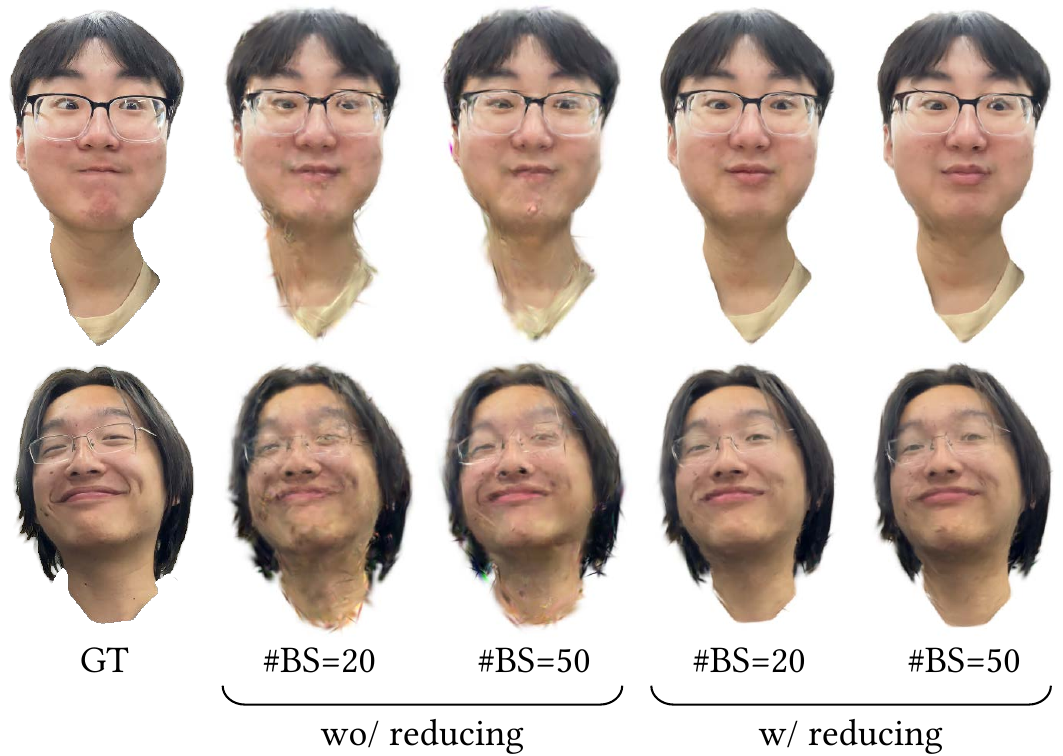}
    \caption{\textbf{Ablation study on reduced blendshapes.} The reduced blendshapes significantly improve the quality under novel expression. \texttt{\#BS} indicates the number of blendshape bases. }\vspace*{-2mm}
    \label{fig:abla_adaptive}
\end{figure}

\noindent \textbf{Reduced Blendshapes.} As shown in \cref{fig:abla_adaptive}, we evaluate the effectiveness of reducing blendshapes. In the first setup, we directly use the truncated FLAME expression coefficients as the weights to drive the blendshape model (\textit{wo/ reducing}). In the second setup, we use an MLP to map the full FLAME parameters to reduced blendshape weights (\textit{w/ reducing}). For the self-reenactment task, we observe that the \textit{w/ reducing} setting achieves significantly better results, even with a lesser number of blendshapes.

\begin{table}[t]
\centering
\begin{tabular}{l|ccc}
    \toprule
    Method & PSNR & SSIM & LPIPS \\
    \midrule
    w/o global & 30.26 & 0.9492 & 0.0801 \\
    w/o local & 30.46 & 0.9508 & 0.0788 \\
    Ours & \textbf{30.86} & \textbf{0.9537} & \textbf{0.0765} \\
    \bottomrule
\end{tabular}
\caption{\textbf{Ablation study on online reconstruction sampling strategy.} Experiments are conducted on INSTA dataset.}
\label{tab:abla_online}
\end{table} 
\begin{table}[t]
\centering
\begin{tabular}{l|ccc}
    \toprule
    Method & Throughput & GPU Utilization \\
    \midrule
    w/o batch parallel & 151 & 52\% \\
    naive batch parallel & 463 & 94\% \\
    Ours & \textbf{630} & \textbf{100\%} \\
    \bottomrule
\end{tabular}
\caption{\textbf{Ablation study on training scheme.}}
\label{tab:abla_training}
\end{table}

\noindent \textbf{Number of Blendshapes.} In \cref{fig:bs_num}, we examine the effect of the number of reduced blendshapes on reconstruction quality and training efficiency. Our model achieves nearly optimal quality with 20 blendshapes, offering a balance between quality and efficiency as training time increases linearly with the number of blendshapes. Compared to GaussianBlendshapes \cite{ma20243d}, our model performs better at any given number of blendshapes, whereas GaussianBlendshapes requires around 80 blendshapes to reach maximum quality.


\noindent \textbf{Batch-Parallel Rasterization.} In \cref{tab:abla_training}, we assess training throughput and GPU utilization under different schemes. Throughput is defined as frames processed
per second. With a naive batch-parallel approach, throughput increases by $3\times$, and with our optimized batch-parallel rasterization process, throughput further improves by $4.2\times$, reaching 100\% GPU utilization.

\noindent \textbf{Online Reconstruction Strategy.} In \cref{tab:abla_online}, we compare our online training strategy (\textit{Ours}) to two variants: one without a global sampling pool (\textit{w/o global}) and one without a local sampling pool (\textit{w/o local}). Our method outperforms both alternative settings.

\begin{figure}[t]
    \centering
    
    \includegraphics[width=1.0\linewidth]{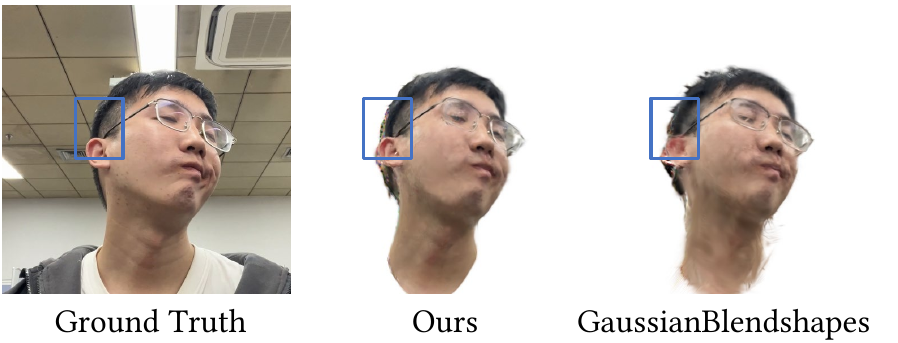}
  
    \caption{\textbf{Limitations.} Our approach produces suboptimal results under large novel-pose.}\vspace*{-2mm}
    \label{fig:limitation}
\end{figure}
 
\section{Limitation and Discussion}
As shown in \cref{fig:limitation}, our constructed avatar models can exhibit apparent artifacts in large novel-pose rendering that are far from training poses, a common problem for head avatar reconstruction methods such as GaussianBlendshapes \cite{ma20243d}. It is worth noting that there is a risk of misuse of our method (e.g., the so-called DeepFakes). We strongly oppose applying our work to produce fake images or videos of individuals with the intention of spreading false information or damaging their reputations.

\section{Conclusion}
We introduce Reduced Gaussian Blendshapes for real-time photorealistic animated avatar reconstruction from monocular video streams. Our blendshape representation is adaptive to the subject, not aligned with 3DMM blendshapes, and has better reconstruction quality with fewer blendshapes. Besides, we introduce a color initialization method and a batch-parallel Gaussian rasterization to speed up the training process, enabling our method to support on-the-fly avatar reconstruction. Experiments show that our method outperforms prior arts both in quality and efficiency.

\section*{Acknowledgments}
This work is supported by NSF China (No. 62172357 \& 62421003) and the XPLORER PRIZE. We thank Bohong Chen and Qing Chang for being our capture subjects.

\clearpage
{
    \small
    \bibliographystyle{ieeenat_fullname}
    \bibliography{main}
}

\clearpage
\setcounter{page}{1}
\maketitlesupplementary


In this supplementary document, we provide additional details about our method, experiments, and results. In \cref{sec:add_method}, we elaborate on the Gaussian transformation calculations, the online frame processing algorithm, and implementation details. \Cref{sec:add_abla} includes further ablation studies, experiments and limitations. Finally, in \cref{sec:add_results}, we showcase online reconstruction results and provide additional qualitative comparisons. We also encourage viewing our supplemental video for a comprehensive demonstration, including a real-time online reconstruction showcase.

\section{Method Details}
\label{sec:add_method}

\subsection{Gaussian Transformation}
\begin{figure}[t]
    \centering
    \includegraphics[width=1.0\linewidth]{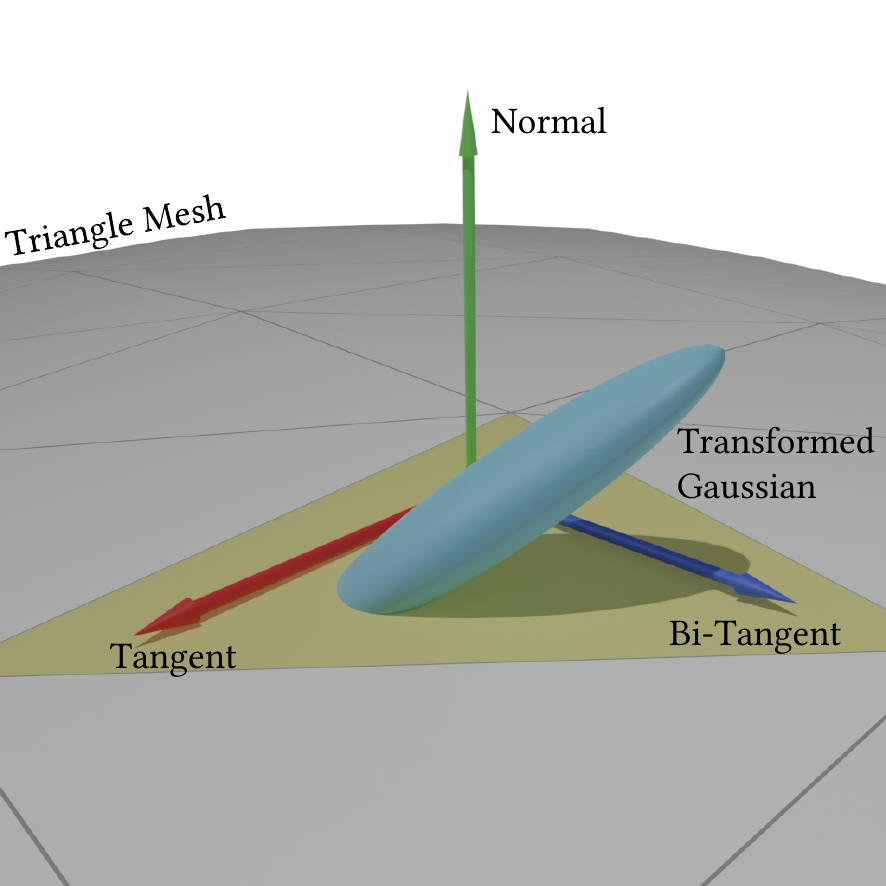}
    \caption{\textbf{Gaussian Transformation.} The corresponding triangle face and the transformed Gaussian are illustrated in yellow and sky blue, respectively. }
    \label{fig:transform}
\end{figure}

To incorporate geometric priors from the 3DMM mesh, we define Gaussian attributes in tangent space $\mathcal{T}$. For rendering, these Gaussians are transformed into deformed space $\mathcal{D}$ based on mesh deformation, as illustrated in \cref{fig:transform}. For a mesh $M^{\theta}$ under expression parameters $\theta$, the transformation involves computing the TBN matrix $\mathbf{R}$ for each triangle (see \cref{alg:tbn}). Gaussian position $\mathbf{x}^{\mathcal{T}}$ and rotation $\mathbf{q}^{\mathcal{T}}$ are transformed to $\mathcal{D}$ as follows:

\begin{equation}
\begin{aligned}
\mathbf{x}^{\mathcal{D}} = \mathbf{R} \mathbf{x}^{\mathcal{T}} + \mathbf{t}, \ \mathbf{q}^{\mathcal{D}} = \mathbf{R} \mathbf{q}^{\mathcal{T}},
\end{aligned}
\end{equation}

where $\mathbf{t}$ is the translation vector, computed via barycentric interpolation using the triangle vertices and the Gaussian's UV coordinates. The UV coordinates are initialized by sampling from the UV map and remain fixed during optimization.

\subsection{Online Frame Processing}
For online reconstruction, each incoming frame is processed sequentially, as shown in \cref{alg:input}. We first use a real-time 3DMM tracker \cite{cao2014displaced} to extract 3DMM parameters $\theta$ from the incoming frame $I_i$ and compute the corresponding mesh $M_i$. The data sample $D_i = {I_i, \theta_i, M_i}$ is then assembled.

If the local sampling pool $\mathcal{M}_l$ is full, the last sample $D_j$ is removed and appended to the global sampling pool $\mathcal{M}_g$ with probability $\frac{|\mathcal{M}_g|}{j}$. Finally, $D_i$ is added to $\mathcal{M}_l$. The local pool $\mathcal{M}_l$ operates in a FIFO manner, while the global pool $\mathcal{M}_g$ uses Reservoir Sampling \cite{vitter1985random}. The avatar model is optimized using batched data sampled from both pools.

\subsection{Implementation Details}
Our method is implemented in PyTorch, with the Gaussian Transformation, Linear Blending, and Color Initialization modules written in CUDA as PyTorch extensions. The batch-parallel Gaussian rasterizer builds upon the 3DGS renderer \cite{kerbl20233d}.

We use the Adam optimizer with a batch size of 10. Learning rates for $\mathbf{x}$, $\alpha$, $\mathbf{s}$, $\mathbf{q}$, and $\mathbf{c}$ are set to $0.0008$, $0.25$, $0.025$, $0.005$, and $0.0125$, respectively. For blendshape parameters $\Delta \mathbf{x}$, $\Delta \mathbf{q}$, and $\Delta \mathbf{c}$, the learning rates are scaled by $0.05$, $0.5$, and $0.5$, relative to $\mathbf{x}$, $\mathbf{q}$, and $\mathbf{c}$. Blendshapes are not applied to Gaussian opacity or scaling. We only use RGB color for appearance modeling. Besides, we apply activation functions after the linear blending to ensure the validity of Gaussian parameters, following GaussianBlendshapes \cite{ma20243d}.

For the offline setting, we optimize our model for 5000 steps. In the online setting, to enable a more reasonable comparison with the offline setting to evaluate our online quality, we simulate a real-time on-the-fly setting by streaming pre-processed video frames and FLAME parameters at 25 FPS. The training steps depend on the length of the video stream, \ie 2-minute video requires about 75k steps (120s$\times$630 training steps/s). We also implement a real-time reconstruction pipeline using a real-time tracker \cite{cao2014displaced} and the FaceWarehouse 3DMM model \cite{cao2013facewarehouse}. All experiments were conducted on a single NVIDIA RTX 3090.

The MLP $\mathcal{F}$ consists of three layers with a 128-dimensional latent feature and is optimized with a learning rate of $0.001$. Following GaussianAvatars \cite{qian2024gaussianavatars}, we add 60 additional triangle faces to model teeth. The UV map resolution is initialized at 256, resulting in approximately 60k Gaussians.

\begin{algorithm}[t]
\caption{TBN Matrix Calculation}
\label{alg:tbn}
\begin{algorithmic}
\Require Triangle vertices $\mathbf{v_0}, \mathbf{v_1}, \mathbf{v_2}$ and texture coordinates $\mathbf{uv_0}=(u_0, v_0), \mathbf{uv_1}=(u_1, v_1), \mathbf{uv_2}=(u_2, v_2)$
\Ensure TBN matrix $\mathbf{R}$

\State $\mathbf{M} \gets \begin{bmatrix}
u_1 - u_0 & u_2 - u_0 \\
v_1 - v_0 & v_2 - v_0
\end{bmatrix}$

\State $\mathbf{e_1} \gets \mathbf{v_1} - \mathbf{v_0}$
\State $\mathbf{e_2} \gets \mathbf{v_2} - \mathbf{v_0}$

\State $\mathbf N \gets \frac {\mathbf{e_1} \times \mathbf{e_2}} {||\mathbf{e_1} \times \mathbf{e_2}||}$

\State $\begin{bmatrix}\mathbf{T} \\ \mathbf{B}\end{bmatrix} \gets \mathbf{M^{-1}} \begin{bmatrix}\mathbf{e_1} \\ \mathbf{e_2}\end{bmatrix}$

\State $\mathbf{R} \gets \begin{bmatrix}\mathbf{T} & \mathbf{B} & \mathbf{N}\end{bmatrix}$

\end{algorithmic}
\end{algorithm}

\begin{algorithm}[t]
\caption{Processing Incoming Frame}
\label{alg:input}
\begin{algorithmic}
\Require Input frame $I_i$, local sampling pool $\mathcal M_l$, global sampling pool $\mathcal M_g$
\Ensure Updated $\mathcal M_l$ and $\mathcal M_g$
\State $\theta_i \gets \mathrm{tracker}(I_i)$
\State $M^{\theta} \gets \mathrm{compute\_3DMM\_mesh} (\theta_i) $
\State $D_i \gets \{ I_i, \theta_i, M^{\theta} \}$
\If{$\mathcal M_l$ is full}
    \State $D_j \gets \mathcal M_l.\mathrm{pop\_front}()$ 
    \If{$\mathcal M_g$ is full} \Comment{Reservoir Sampling}
        \State $k \gets \mathrm{randint}(0, j)$
        \If{$k < |M_g|$} \Comment{$|\mathcal M_g|$ is the size of $\mathcal M_g$}
            \State $\mathcal M_g[k] \gets D_j$
        \EndIf
    \Else
        \State $\mathcal M_g.\mathrm{append}(D_j)$
    \EndIf
\EndIf
\State $\mathcal M_l.\mathrm{append}(D_i)$
\end{algorithmic}
\end{algorithm}

\begin{figure}[t]
    \centering
    \includegraphics[width=1.0\linewidth]{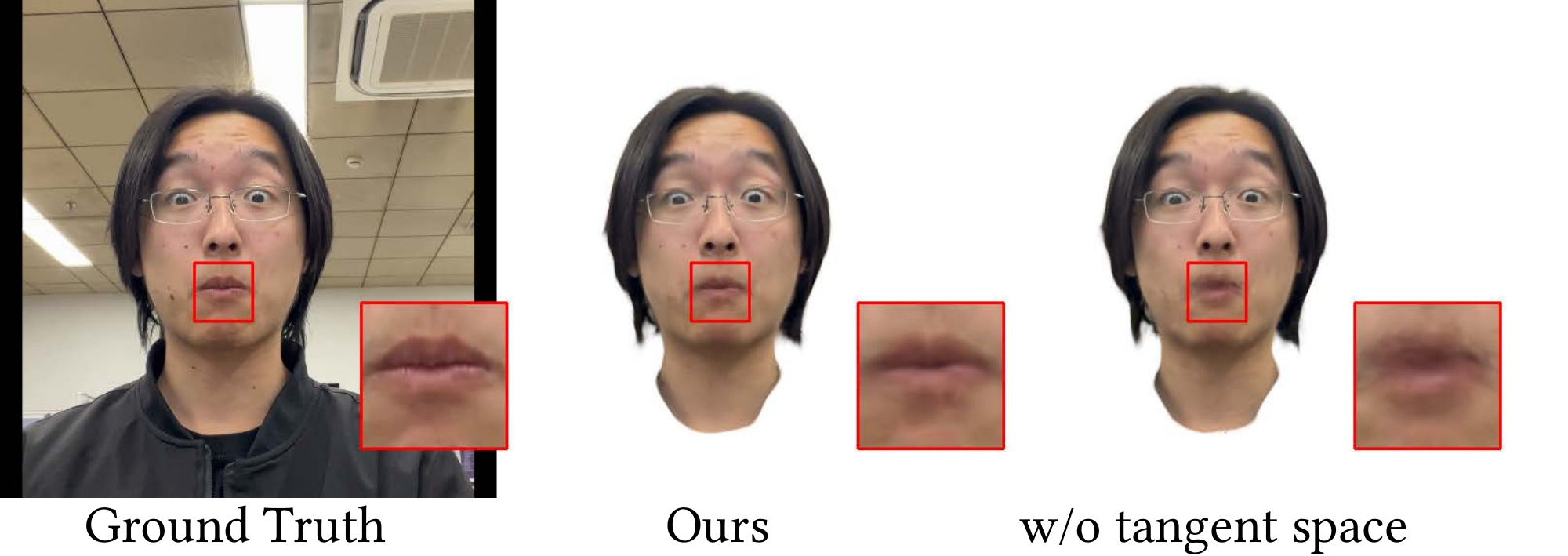}
    \caption{\textbf{Impact of tangent space.} Without geometry prior provided by mesh deformation, the model delivers blurred results.}
    \label{fig:abla_tangent}
\end{figure}
\begin{figure}[t]
    \centering
    \includegraphics[width=1.0\linewidth]{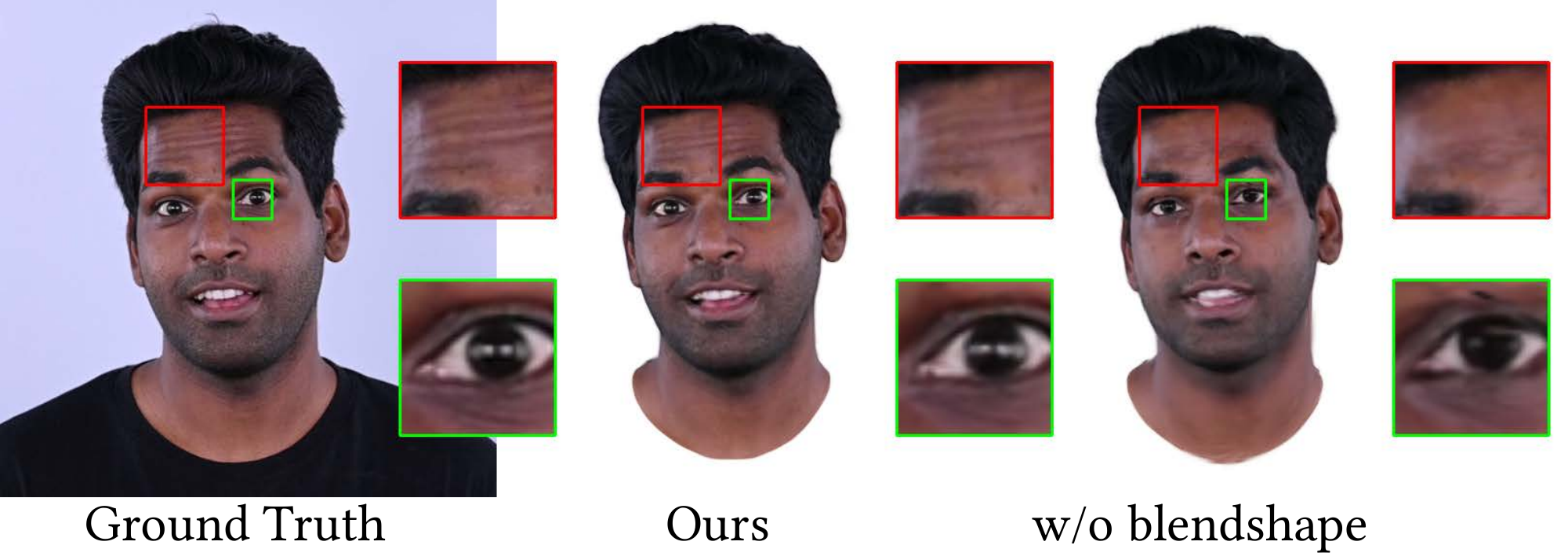}
    \caption{\textbf{Impact of blendshape.} Without blendshapes, the model lacks representation capacity to model wrinkle-level details. }
    \label{fig:abla_bs}
\end{figure}
\begin{figure}[t]
    \centering
    \includegraphics[width=1.0\linewidth]{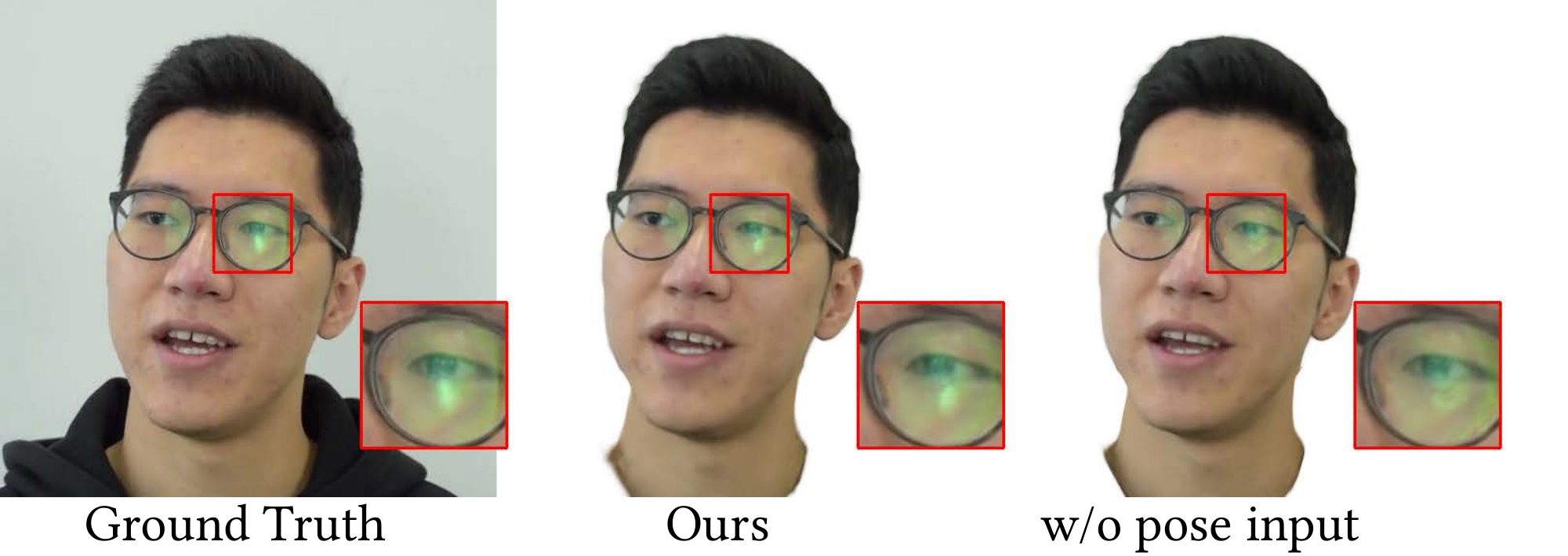}
    \caption{\textbf{Impact of pose input.} Without joint pose input, the model lacks information represents pose-relevant appearance, such as eyeglass reflection. }
    \label{fig:abla_pose}
\end{figure}
\begin{table}[t]
\centering
\begin{tabular}{l|ccc}
    \toprule
    Method & PSNR & SSIM & LPIPS \\
    \midrule
    w/o tangent space & 31.01 & 0.9513 & 0.0825 \\
    w/o blendshape & 29.12 & 0.9402 & 0.0920 \\
    w/o pose input & 30.79 & 0.9533 & 0.796 \\
    Ours (linear) & \textbf{31.38} & 0.9556 & 0.772 \\
    Ours & 31.19 & \textbf{0.9565} & \textbf{0.737} \\
    \bottomrule
\end{tabular}
\caption{\textbf{Ablation study on model design.} Our default setting outperforms other variants. Experiments are conducted on INSTA dataset. }
\label{tab:abla_model}
\end{table}

\begin{table*}[t]
    \centering
    \begin{tabular}{l|cccc}
        \toprule
        Method & PSNR$\uparrow$ & SSIM$\uparrow$ & LPIPS$\downarrow$ & $e_{orth}$ \\
        \midrule
        Orthogonal loss & 30.91 & 0.9499 & 0.0890 & $9.164 \times 10^{-4}$ \\
        QR decomposition & 30.26 & 0.9462 & 0.0912 & $9.558 \times 10^{-8}$ \\
        Schmidt orthogonalization & 30.75 & 0.9492 & 0.0892 & $4.157 \times 10^{-8}$ \\
        Ours (not orthogonal) & \textbf{31.16} & \textbf{0.9551} & \textbf{0.0725} & 3.173 \\
        \bottomrule
    \end{tabular}
    \caption{\textbf{Investigation of basis orthogonality}. Experiments are conducted on INSTA dataset.}
    \label{tab:orthogonal}
\end{table*}

\section{Additional Experiments}
\label{sec:add_abla}

\subsection{Ablation on Model Design}
We conducted ablation studies to evaluate our model design, with quantitative results shown in \cref{tab:abla_model}. Our default design consistently outperforms other variants.

First, we tested a variant without defining Gaussians in tangent space (\textit{w/o tangent space}), where Gaussians are transformed using joint poses instead of mesh deformation, similar to GaussianBlendShapes \cite{ma20243d}. Since our reduced blendshapes are subject-adaptive, initializing with generic FLAME blendshapes is infeasible. Consequently, the joint pose rotations provide insufficient geometric priors, leading to suboptimal Gaussian transformations. As shown in \cref{fig:abla_tangent}, this variant relies on color composition to fit subtle deformations (e.g., around the mouth), resulting in blurred self-reenactment results.

Second, we evaluated a variant without blendshapes (\textit{w/o blendshape}), where only a single set of base Gaussians is attached to the 3DMM mesh, similar to SplattingAvatar \cite{shao2024splattingavatar} and GaussianAvatars \cite{qian2024gaussianavatars}. This configuration lacks the capacity to model finer details, such as wrinkles or eyeball reflections, as shown in \cref{fig:abla_bs}.

Third, we ablated the FLAME parameters provided to the MLP, which are mapped to reduced blendshape weights. Our default setting (\textit{Ours}) includes both expression coefficients and joint poses. When joint poses are excluded (\textit{w/o pose input}), the model fails to represent pose-dependent appearance features, such as eyeglass reflections (\cref{fig:abla_pose}).

Finally, we replaced the MLP in our model with a simple linear projection (\textit{Ours (linear)}). As shown in \cref{tab:abla_model}, while this variant achieves slightly higher PSNR, it performs worse on SSIM and LPIPS metrics.

\subsection{Ablation on Online Reconstruction Strategy}
We further analyzed the importance of our global and local sampling strategy for online reconstruction.

\begin{figure}[t]
    \centering
    \includegraphics[width=1.0\linewidth]{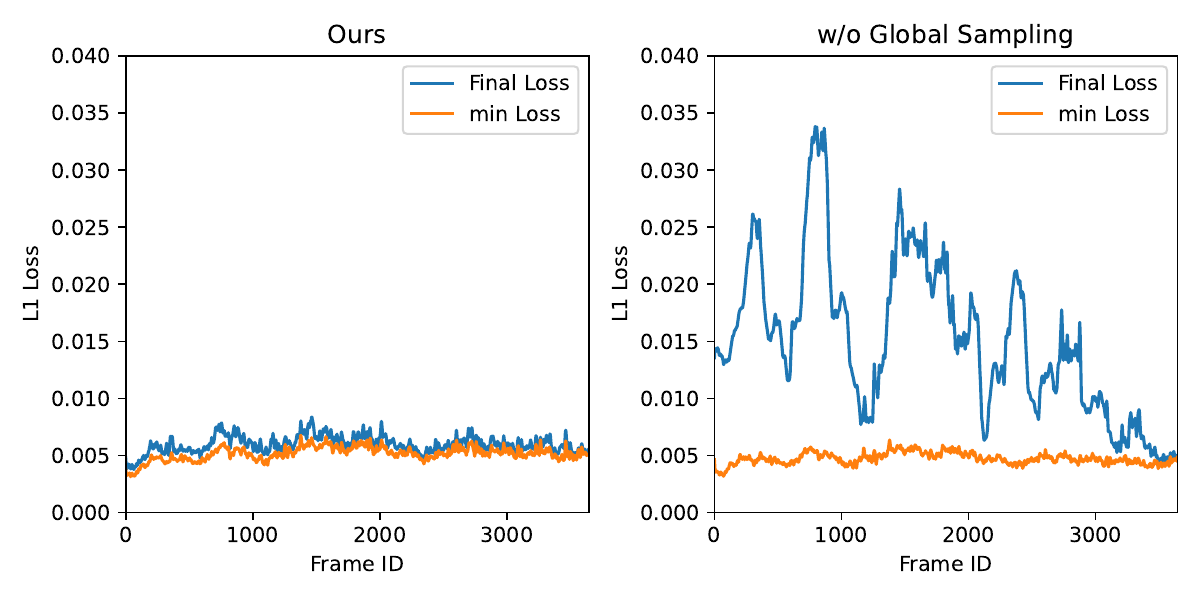}
    \caption{\textbf{Impact of global sampling on online reconstruction.} Left: per frame L1 loss with our sampling strategy. Right: per frame L1 loss without gloabal sampling. }
    \label{fig:abla_global}
\end{figure}

\cref{fig:abla_global} compares the per-frame L1 loss with (left) and without (right) global sampling. The blue line shows the L1 loss after online training, while the orange line indicates the minimum L1 loss during training. The gap between these lines reflects the extent of "forgetting." Our sampling strategy effectively mitigates this issue.

Additionally, we examined the impact of pool size on reconstruction quality (\cref{tab:abla_online_pool}). The results show that varying pool sizes has minimal effect. Based on empirical observations, we set the local pool size to $|\mathcal{M}_l| = 150$ and the global pool size to $|\mathcal{M}_g| = 1000$.

 \begin{figure}[t]
    \centering
    \includegraphics[width=1.0\linewidth]{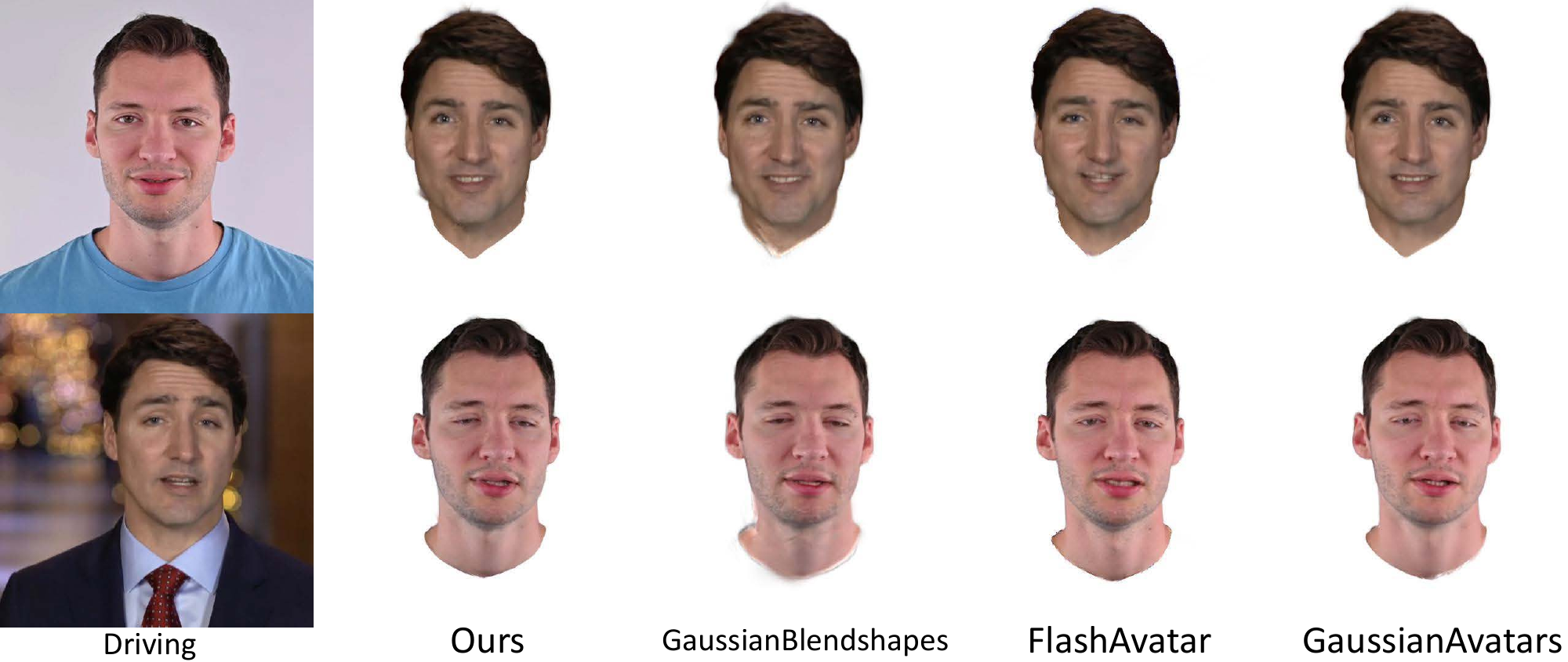}
    \caption{\textbf{Qualitative comparisons of cross-id reenactment.}}
    \label{fig:reenact_compare}
\end{figure}
\begin{table}[t]
\centering
\begin{tabular}{l|ccc}
    \toprule
    Method & PSNR & SSIM & LPIPS \\
    \midrule
    $|\mathcal M_l| = 150, |\mathcal M_l| = 500$ 
    & 30.90 & 0.9537 & 0.0758 \\
    $|\mathcal M_l| = 150, |\mathcal M_l| = 1000$ 
    & 31.04 & 0.9543 & 0.0758 \\
    $|\mathcal M_l| = 150, |\mathcal M_l| = 1500$ 
    & 30.80 & 0.9544 & 0.0755 \\
    $|\mathcal M_l| = 150, |\mathcal M_l| = 2000$ 
    & \textbf{31.06} & 0.9546 & 0.0759 \\
    $|\mathcal M_l| = 50, |\mathcal M_l| = 1000$ 
    & 30.94 & 0.9542 & 0.0763 \\
    $|\mathcal M_l| = 100, |\mathcal M_l| = 1000$ 
    & 30.90 & \textbf{0.9548} & 0.0755 \\
    $|\mathcal M_l| = 200, |\mathcal M_l| = 1000$ 
    & 30.97 & 0.9546 & \textbf{0.0754} \\
    \bottomrule
\end{tabular}
\caption{\textbf{Ablation study on sampling pool size of online reconstruction.} Experiments are conducted on INSTA dataset.}
\label{tab:abla_online_pool}
\end{table}
 \begin{table}[t]
    \centering
    \begin{tabular}{l|cccc}
        \toprule
        Method & AED$\downarrow$ & APD$\downarrow$ & CSIM$\uparrow$ \\
        \midrule
        MonoGaussianAvatar & 10.9268 & 0.1406 & 0.7618 \\
        GaussianAvatars & 9.9394 & \underline{0.1167} & 0.7858 \\
        FlashAvatar & 9.9210 & 0.1250 & \textbf{0.7952}  \\
        GaussianBlendShapes & \textbf{9.1241} & \textbf{0.1158} & 0.7866 \\
        Ours & \underline{9.2683} & 0.1179 & \underline{0.7868} \\
        \bottomrule
    \end{tabular}
    \caption{\textbf{Quantitative comparisons of cross-id reenactment.}}
    \label{tab:cross_eval}
\end{table}

\subsection{Investigation of Basis Orthogonality}
The blendshape bases of FLAME are constructed using PCA, making them inherently orthogonal. However, our reduced blendshape bases break this orthogonality during training. Therefore, we examine whether maintaining orthogonality is necessary. First, we formally define the orthogonality of the 20 reduced blendshape bases as
\begin{equation}
    e_{orth} = ||VV^T - I||_2,
\end{equation}
where $V \in \mathbb{R}^{20 \times d}$ represents the concatenation of the normalized 20 bases. The bases are perfectly orthogonal when $e_{orth}=0$, although minor numerical errors may arise in practice. Then we conduct three different experiments to ensure orthogonality: we add 1) an \textit{orthogonal loss} with loss weight $\lambda_{orth} = 10$. we perform 2) \textit{QR decomposition} or 3) \textit{Schmidt orthogonalization}, both every 1k training steps. As shown in \cref{tab:orthogonal}, our method achieves the best result. Since our objective is to reconstruct a photorealistic Gaussian avatar, it is not necessary for the blendshape bases to be orthogonal.

\subsection{Comparison of Generalization Ability}
 We futher evaluate 10 cross-identity reenactment video sequences using standard metrics from the image animation literature \cite{qing2024enhancing}: average expression distance(AED), average pose distance(APD), and identity similarity(CSIM). As shown in \cref{tab:cross_eval} and \cref{fig:reenact_compare}, our method yields comparable results to other Gaussian-based methods.

\begin{figure}[t]
    \centering
    \includegraphics[width=1.0\linewidth]{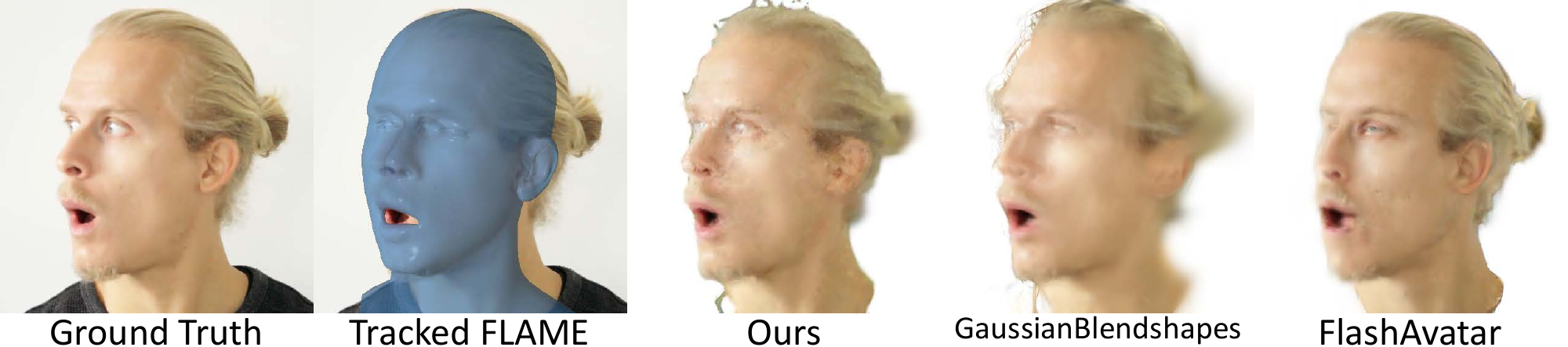}
    \caption{\textbf{Limitation of tracking error.}}
    \label{fig:tracking}
\end{figure}

 \subsection{Limitation of Tracking Error}
 Our method, along with others, exhibits artifacts caused by tracking errors. We demonstrate a common tracking error that occurs with a large side head pose in \cref{fig:tracking}.

\section{Additional Results}
\label{sec:add_results}
\begin{figure}[t]
    \centering
    \includegraphics[width=1.0\linewidth]{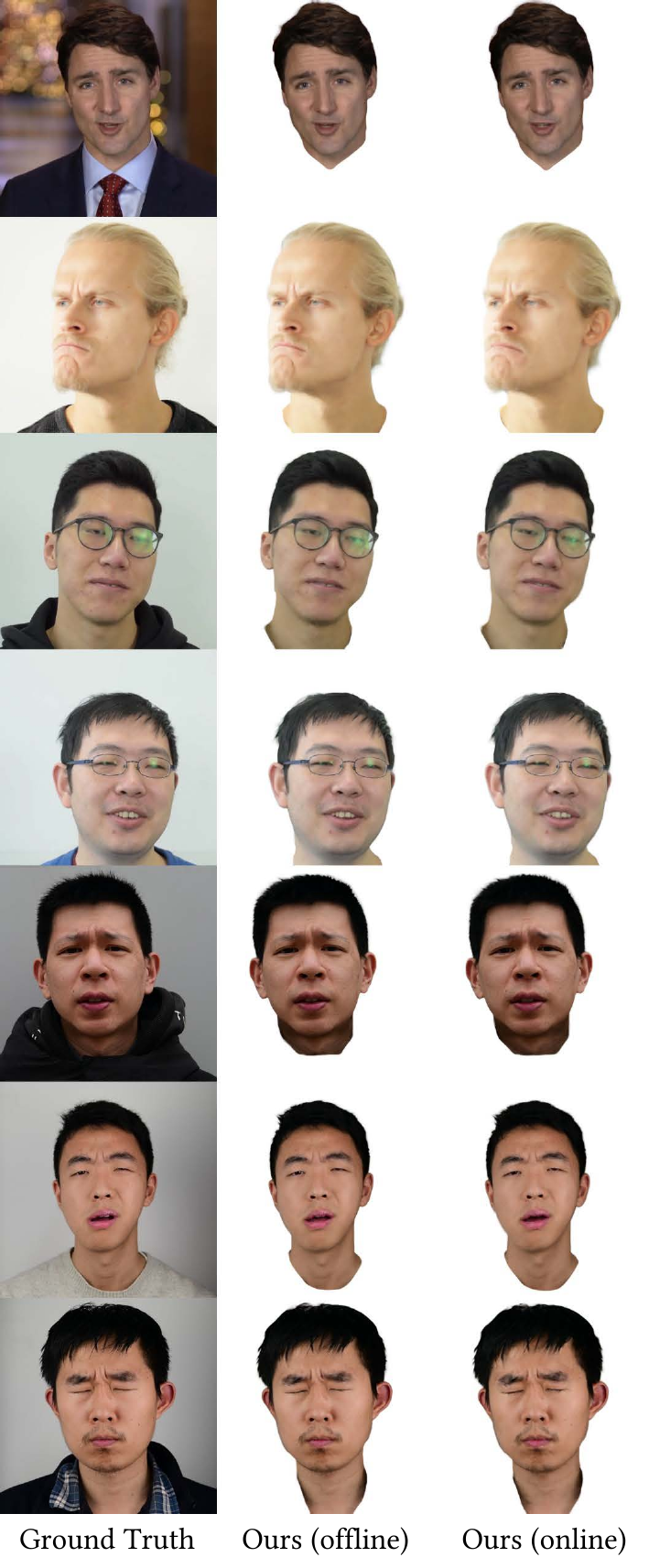}
    \caption{\textbf{Results of online reconstruction.} Our online reconstruction strategy achieves comparable quality to offline setting. }
    \label{fig:online}
\end{figure}

\begin{figure*}[t]
    \centering
    \includegraphics[width=1.0\linewidth]{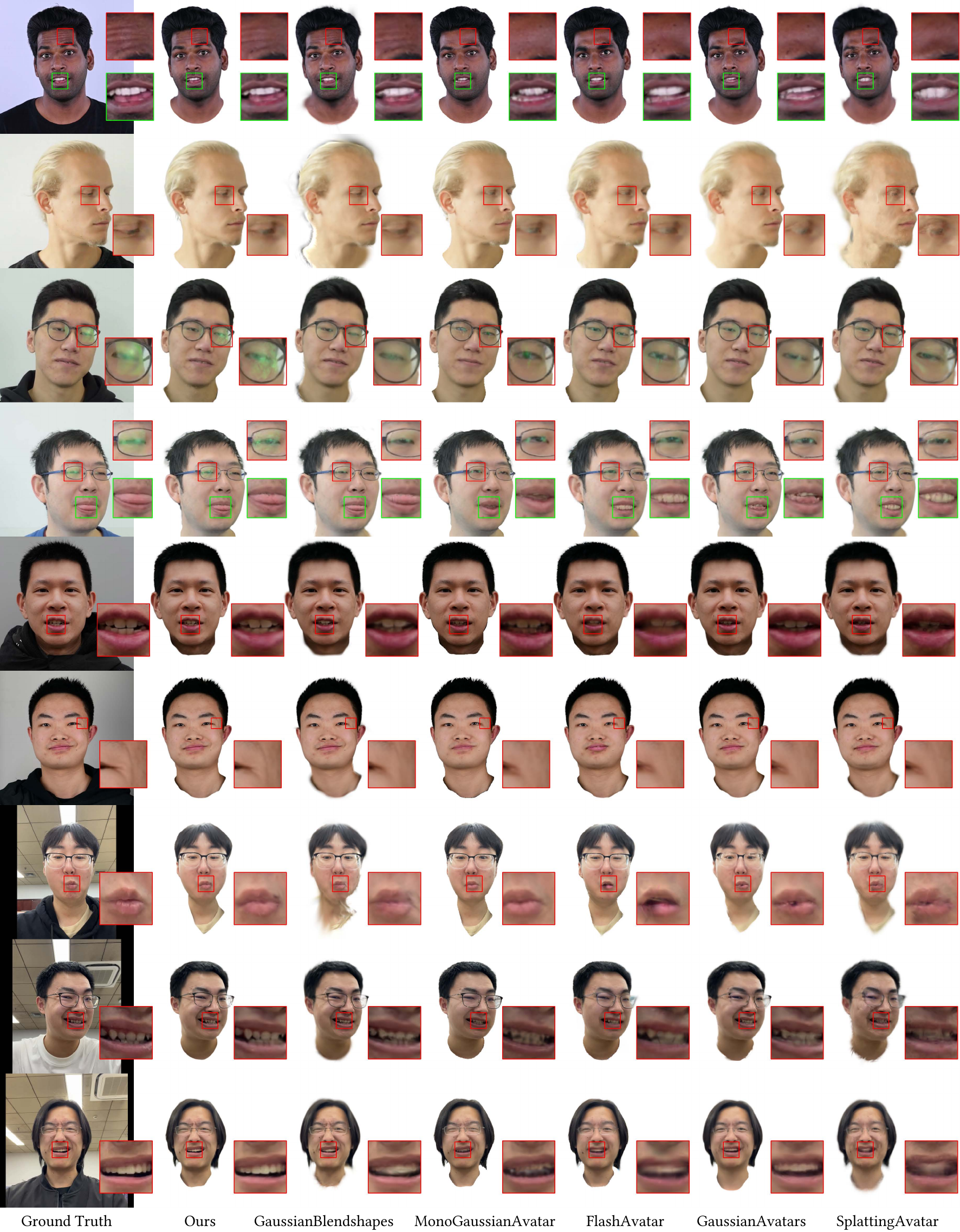}
    \caption{\textbf{Additional Qualitative comparisons.}}
    \label{fig:more_compare}
\end{figure*}

We present the online reconstruction results in \cref{fig:online}, demonstrating comparable quality to the offline setting. Additional qualitative comparisons with other methods are provided in \cref{fig:more_compare}.


\end{document}